%% file: main.tex
\documentclass[runningheads]{llncs}

\usepackage{amsmath,amssymb} 
\usepackage{color}
\usepackage[width=122mm,left=12mm,paperwidth=146mm,height=193mm,top=12mm,paperheight=217mm]{geometry}

\usepackage[boxed]{algorithm2e}
\usepackage{caption}
\usepackage{mathtools}
\usepackage{amsfonts}
\usepackage{booktabs}
\usepackage{graphicx}
\usepackage{soul}
\usepackage{color}
\usepackage{xspace}
\usepackage[subrefformat=parens]{subfig}
\usepackage{soul}

\usepackage{amsfonts}
\usepackage{graphicx}
\usepackage{multirow}
\usepackage{bm}
\usepackage{color}
\usepackage{tabularx}
\usepackage{footnote}
\usepackage{threeparttable}
\usepackage{tablefootnote}
\usepackage{fancyhdr}
\usepackage{amssymb}
\usepackage{pifont}

\usepackage{wrapfig}
\usepackage{setspace}
\usepackage{xcolor}         

\usepackage{exscale}
\usepackage{relsize}
\usepackage{xr}

\usepackage{mcode}

\usepackage{colortbl}
\usepackage{wrapfig}

\definecolor{cyan}{cmyk}{.3,0,0,0}
\definecolor{lightblue}{HTML}{B0C4DE}
\definecolor{darkblue}{HTML}{177CB0}
\definecolor{maroon}{cmyk}{0,0.87,0.68,0.32}

\usepackage{caption}

\newcommand{\pz}[1]{#1}

\newcommand{\eg}{e.g.}
\newcommand{\ie}{i.e.}

\input{MACPdef_manu.tex}

\newcommand\blfootnote[1]{%
	\begingroup
	\renewcommand\thefootnote{}\footnote{#1}%
	\addtocounter{footnote}{-1}%
	\endgroup
}

\newlength\savewidth\newcommand\shline{\noalign{\global\savewidth\arrayrulewidth 
		\global\arrayrulewidth 1pt}\hline\noalign{\global\arrayrulewidth\savewidth}}   
\newcommand{\tablestyle}[2]{\setlength{\tabcolsep}{#1}\renewcommand{\arraystretch}{#2}\centering\footnotesize}

\begin{document}
\pagestyle{headings}
\mainmatter
\def\ECCVSubNumber{2990}  

\title{Mugs: A Multi-Granular Self-Supervised  Learning Framework} 

\titlerunning{Mugs: A Multi-Granular Self-Supervised  Learning Framework}

\authorrunning{Preprint}  

\author{\normalsize{Pan Zhou$^{1*}$}   \  \normalsize{Yichen Zhou$^{1*}$}   \   \normalsize{Chenyang Si$^{1*}$}    \    \normalsize{Weihao Yu$^{1}$}    \   \normalsize{Teck Khim Ng$^{2}$}   \    \normalsize{Shuicheng Yan$^{1}$}\\
	\institute{$^{1}$ Sea AI Lab, Singapore \qquad  $^{2}$ National University of Singapore} 
}
\maketitle

\begin{abstract}
In self-supervised learning\blfootnote{$^{*}$Equal contribution. Emails:  \{zhoupan, zhouyc, sicy, yuweihao, Yansc\}@sea.com,  ngtk@comp.nus.edu.sg},  multi-granular features  are  heavily desired though  rarely investigated, as different downstream tasks (\eg, general and fine-grained classification)  often require different or multi-granular features, \eg~fine- or coarse-grained one or  their mixture.  
In this work, for the first time, we propose an effective MUlti-Granular Self-supervised  learning (Mugs) framework to explicitly learn  multi-granular visual features. 
Mugs has three complementary granular supervisions:  
1)\! an instance discrimination supervision (IDS), 
2)\! a novel  local-group discrimination supervision (LGDS), 
and 3)\! a group discrimination supervision (GDS).  
IDS distinguishes different instances to learn  instance-level fine-grained features.
LGDS aggregates features of an image and its neighbors into a local-group feature, and pulls local-group features from different crops of the same image together and push them away for others.  
It provides complementary instance supervision to IDS via an extra alignment on local neighbors,
and scatters different local-groups separately to increase discriminability.  
Accordingly, it helps learn high-level fine-grained features at a local-group level.
Finally, to prevent similar local-groups from being scattered randomly or far away,  GDS brings similar samples close and thus pulls similar local-groups together,  capturing coarse-grained features at a (semantic) group level. Consequently,  Mugs can  capture   three  granular features that  often enjoy higher generality  on diverse downstream tasks over single-granular features, \eg~instance-level fine-grained features in contrastive learning. 
By only pretraining on ImageNet-1K,  Mugs sets new SoTA linear probing accuracy \pz{82.1$\%$} on ImageNet-1K and improves previous  SoTA by  \pz{$1.1\%$}.  	It also surpasses SoTAs on  other tasks,  \eg~transfer learning,  detection  and   segmentation. Codes and  models are available at \textcolor[rgb]{0.925,0,0.549}{\url{https://github.com/sail-sg/mugs}}.

\vspace{-0.7em}
\keywords{self-supervised learning,  multi-granular representation}
\end{abstract}

\vspace{-2.8em}
\section{Introduction}\label{introduction}
\vspace{-0.7em}
The family of self-supervised learning (SSL) approaches~\cite{he2020momentum,chen2020improved,chen2020big,chen2020simple,caron2020unsupervised,grill2020bootstrap,li2020prototypical} aims to learn highly transferable unsupervised representation for various  downstream tasks by training deep models on a large-scale unlabeled dataset.  
To this end, a pretext task, \eg~jigsaw puzzle~\cite{noroozi2016unsupervised} or  orientation~\cite{komodakis2018unsupervised}, is elaborately designed to generate pseudo labels of unlabeled visual data  which are then utilized\\ to train a model without using manual annotations. 
Since unlabeled visual data are of huger amount  and also much cheaper than the manually annotated data,  SSL has been very popularly adopted for visual representation learning recently~\cite{he2020momentum,chen2020simple,caron2020unsupervised,grill2020bootstrap,zhou2021SLR}, and is showing greater potential than supervised learning approaches for learning  highly-qualified and well-transferable representation.

\noindent{\textbf{Motivation.}}  In practice,  various downstream tasks in  SSL field often require   different granular features, such as coarse- or fine-grained features.  
For instance, general classification downstream tasks distinguish a category from other categories and typically desire coarse-grained features, while fine-grained  classification often discriminates subordinate categories and needs more fine-grained features. 
Actually, many downstream tasks highly desire multi-granular features.   
Take the classification task on ImageNet-1K~\cite{ImageNet} as an example.
One needs coarse-grained features to distinguish a big category, \eg~dog,  from other categories, \eg~bird and car, and also requires fine-grained features to discriminate different subordinate categories, such as Labrador and poodle in the dog category. 
However, this important multi-granularity requirement is  ignored in the current  state-of-the-art  SSL  approaches, including the representative  contrastive learning family~\cite{hadsell2006dimensionality,he2020momentum,hjelm2018learning} and clustering learning family~\cite{caron2018deep,DINO}.   For contrastive learning, its instance discrimination task only aims to distinguish individual instances for learning more instance-level fine-grained features, and does not consider the coarse-grained cluster structure in the data. As a result, it cannot well push semantically similar instances to be close either empirically~\cite{zhou2021SLR} or theoretically~\cite{wang2020understanding},  impairing performance, especially for classification downstream tasks. 
The clustering learning family aims to cluster similar instances into the same cluster and thus learns  cluster-level coarse-grained features. In this way, it cannot well handle the downstream tasks that require some fine-grained features.  Therefore, in absence of prior feature preference of down stream tasks, one should build an SSL framework to learn multi-granular representation to well handle as many downstream tasks as possible.

\begin{wrapfigure}{r}{0.332\textwidth}
	\vspace{-1.6em}
	\setlength{\tabcolsep}{0.8pt}  \begin{center}
		\begin{tabular}{c c c}
			\includegraphics[width=0.36\linewidth]{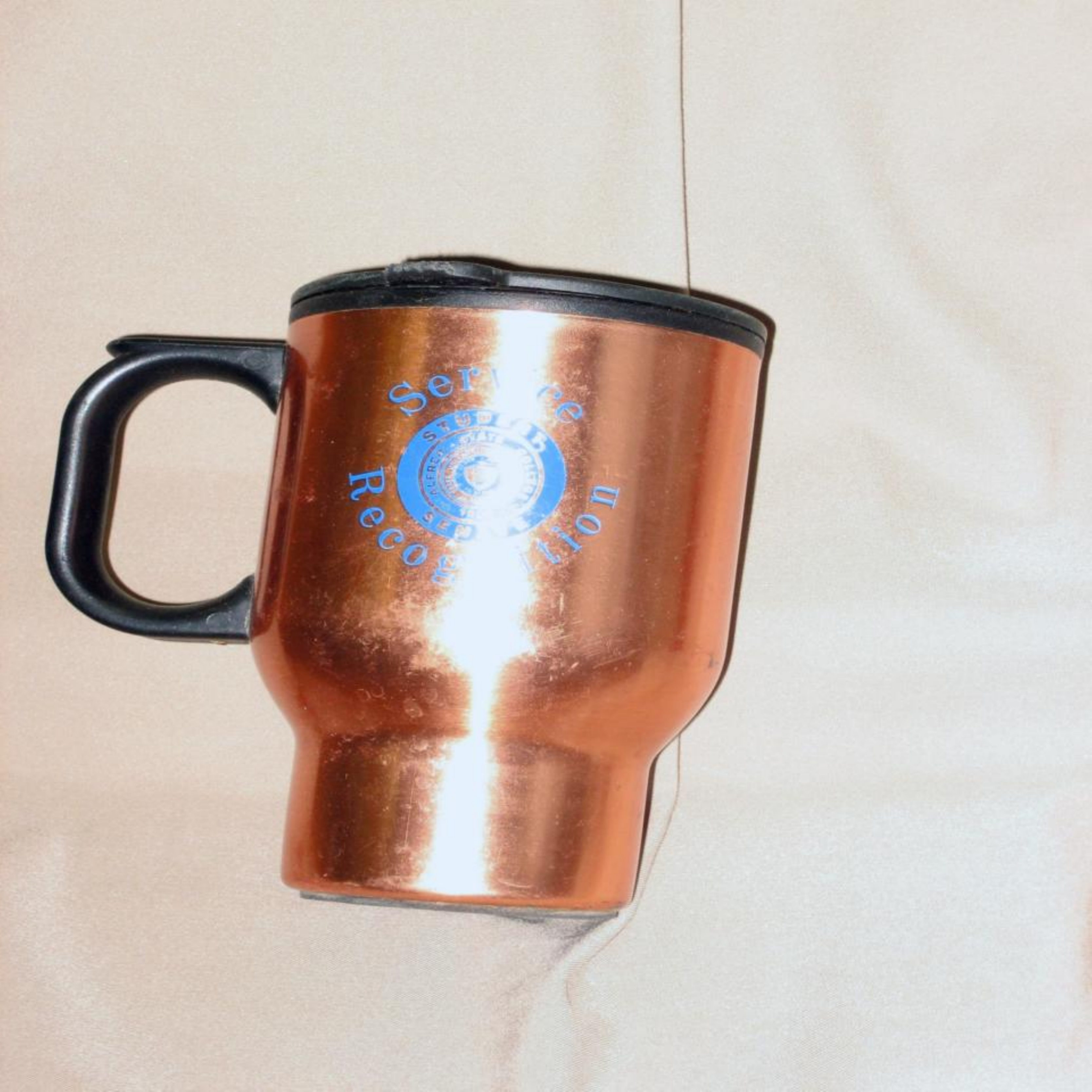} \quad &  \quad \includegraphics[width=0.36\linewidth]{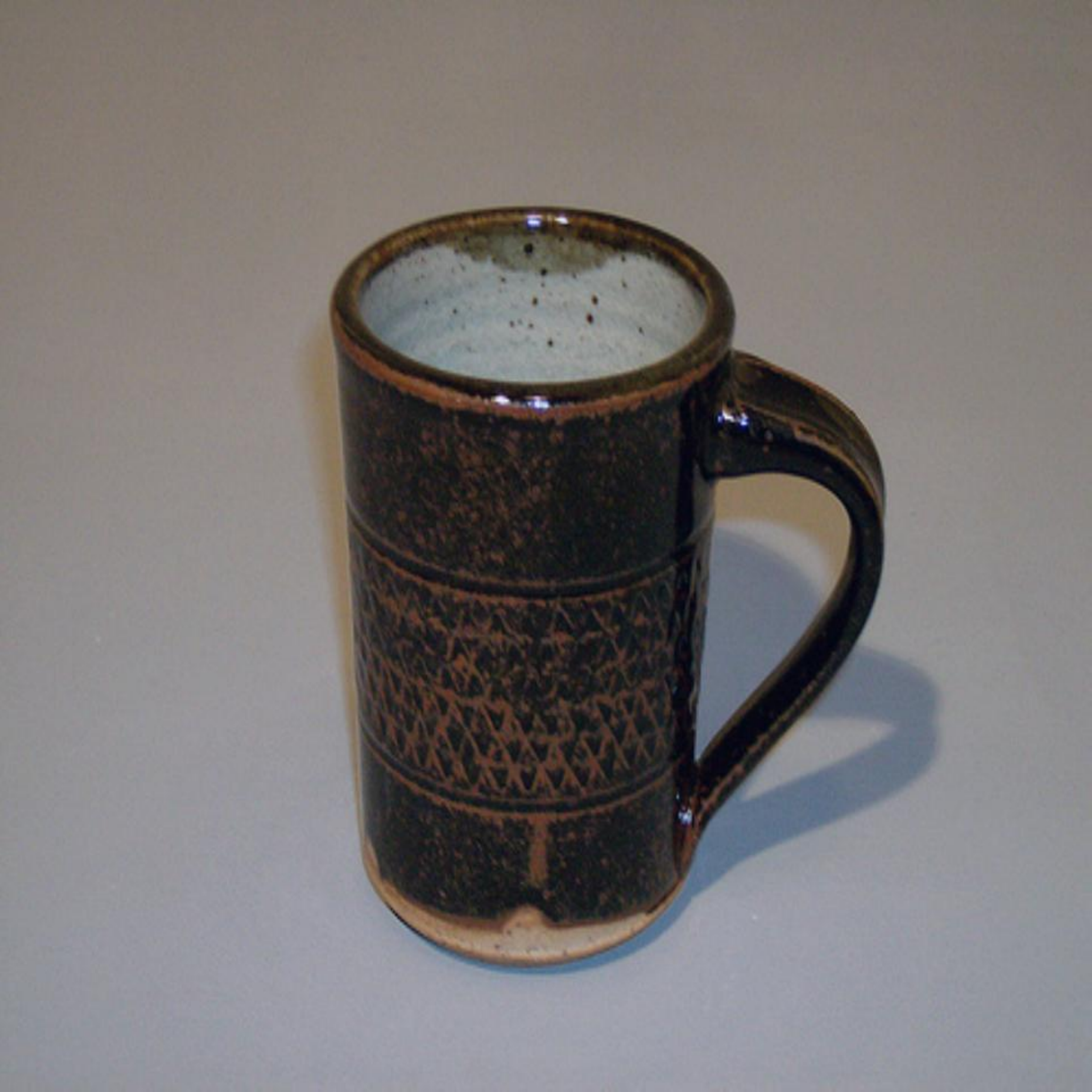} \\ 
			\includegraphics[width=0.36\linewidth]{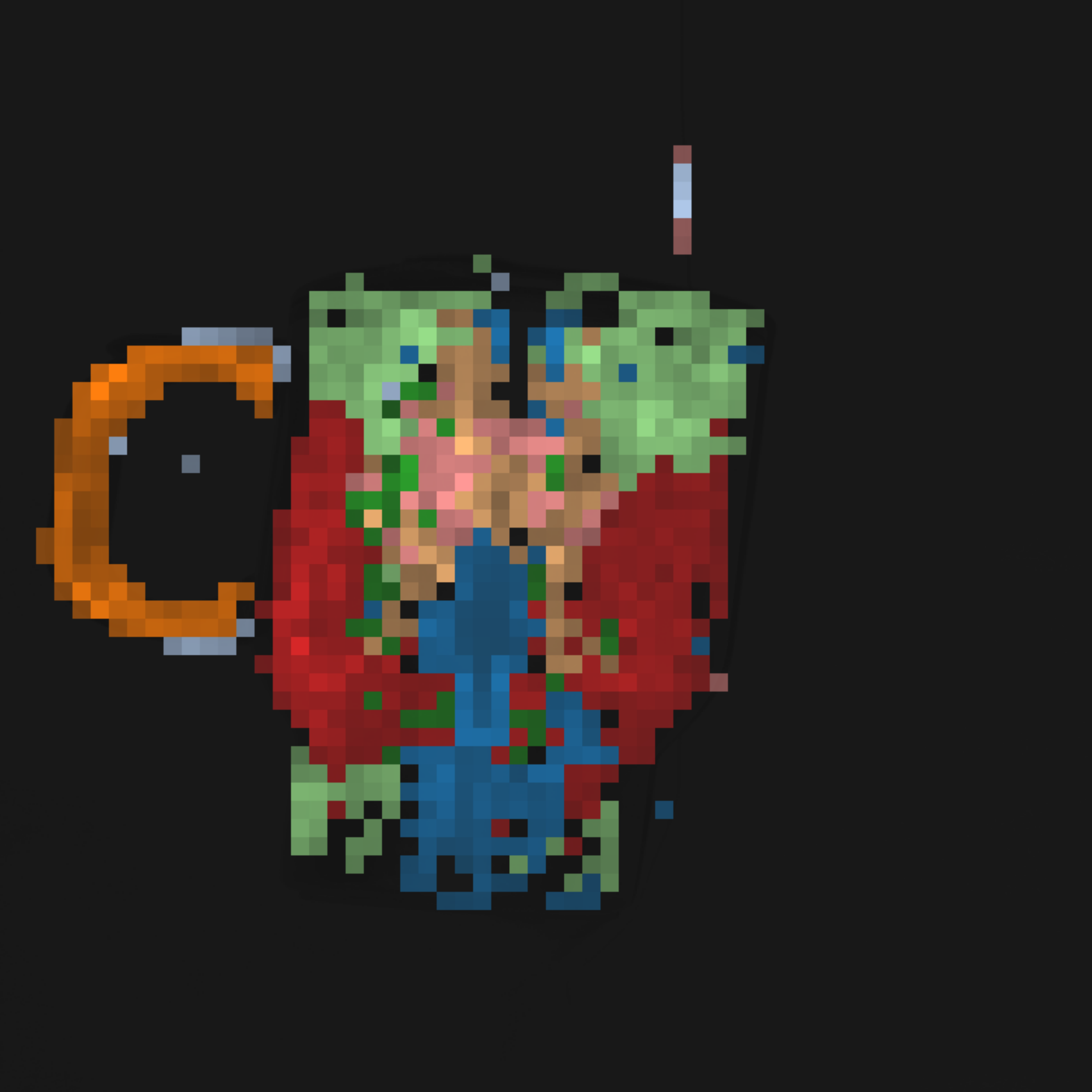} \quad & \quad
			\includegraphics[width=0.36\linewidth]{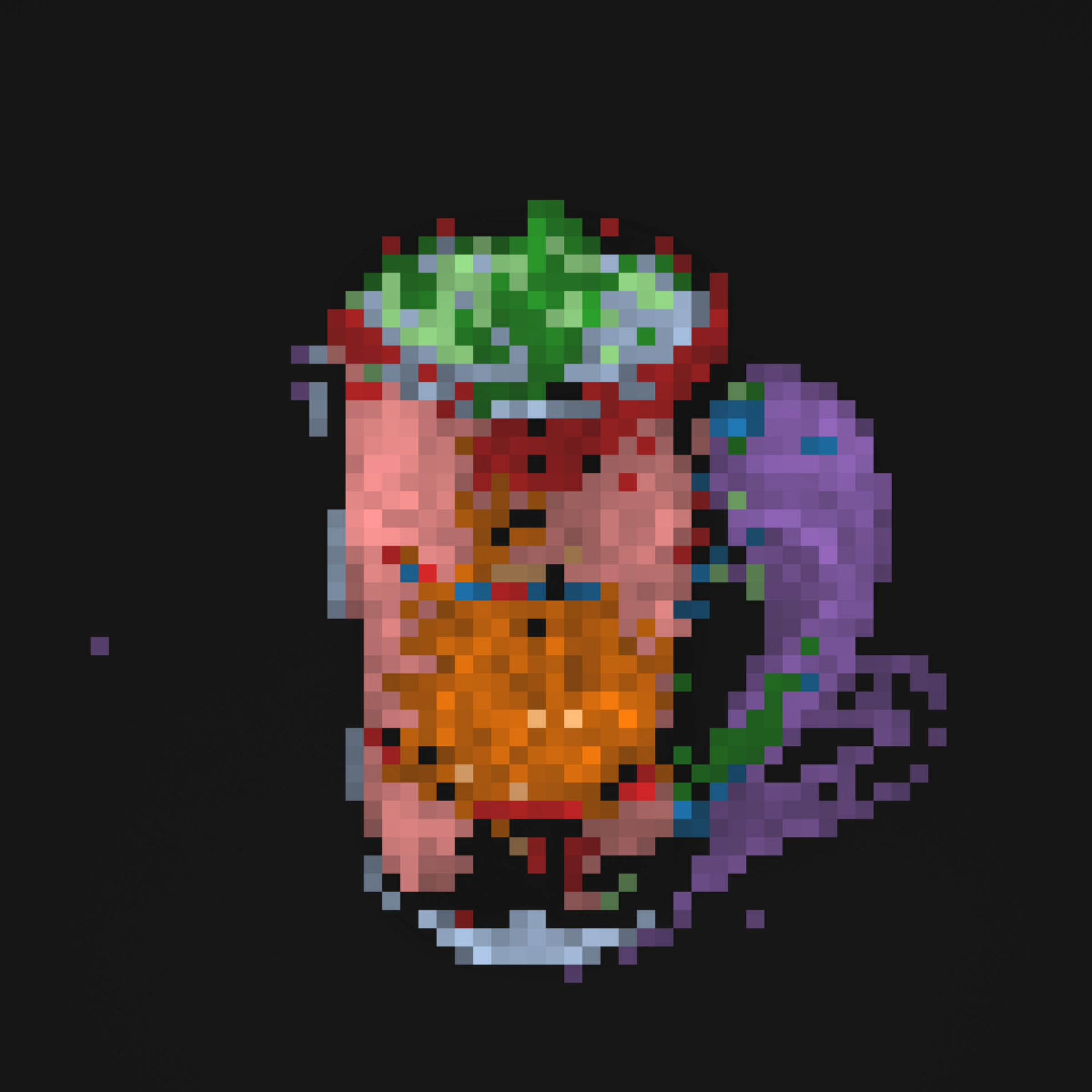} \\
		\end{tabular}  
	\end{center}
	\vspace{-2.2em} 
	\caption{\!\!Attention visualization   on\!  ``mugs"\! of\!   ViT-B\!/\!16 trained   by our  Mugs.  See more examples  in Sec.~\ref{resultsonImagenet}. }
	\label{illustration_attention2}
	\vspace{-0.2em} 
\end{wrapfigure}
\noindent{\textbf{Contributions.}}  
In this work, we propose an effective MUlti-Granular Self-supervised learning (Mugs) framework to explicitly  learn multi-granular features for visual data. 
Specifically, Mugs adopts three complementary granular supervisions:  1) instance discrimination supervision (IDS), 
2) local-group discrimination supervision (LGDS), 
and 3) group discrimination supervision (GDS).  
Inspired by contrastive learning, IDS  distinguishes instances via scattering different instance features separately, and thus supervises instance-level fine-grained feature learning. 
To capture the higher-level fine-grained feature which is also called the ``local-group feature" in this work, Mugs proposes a novel and effective LGDS.
LGDS aggregates the features of an instance and its few highly similar neighbors into a local-group feature through a small transformer.
Then it brings local-group features of  different crops from the same image together and pushes them far away for others.
This supervision enhances Mugs from two aspects: 1) it provides complementary instance supervision to the above IDS, since it enforces different crops of the same image to have highly similar neighbors, which is an extra challenging alignment, and can boost local-group semantic alignment;
2) it encourages highly-similar instances to constitute small local-groups and scatters these groups separately, boosting more discriminative semantic learning. 
Finally, GDS is designed to avoid the cases that similar local-groups are scattered randomly or far away. To this end, GDS brings similar samples together and thus pulls similar local-groups close, which captures coarse-grained features at a (semantic) group level.   
With these complementary supervisions, Mugs can well learn multi-granular data features  which can accurately capture the data semantics, \eg~the shapes of ``mugs" in ImageNet-1K  as illustrated in Fig.~\ref{illustration_attention2}, and also often enjoy better generality and transferability on diverse downstream tasks than single-granular features.

\begin{figure}[tbp]
	\hspace{-1.1em}
	\begin{minipage}[c]{.55\linewidth}
		\centering
		\includegraphics[width=0.81\textwidth]{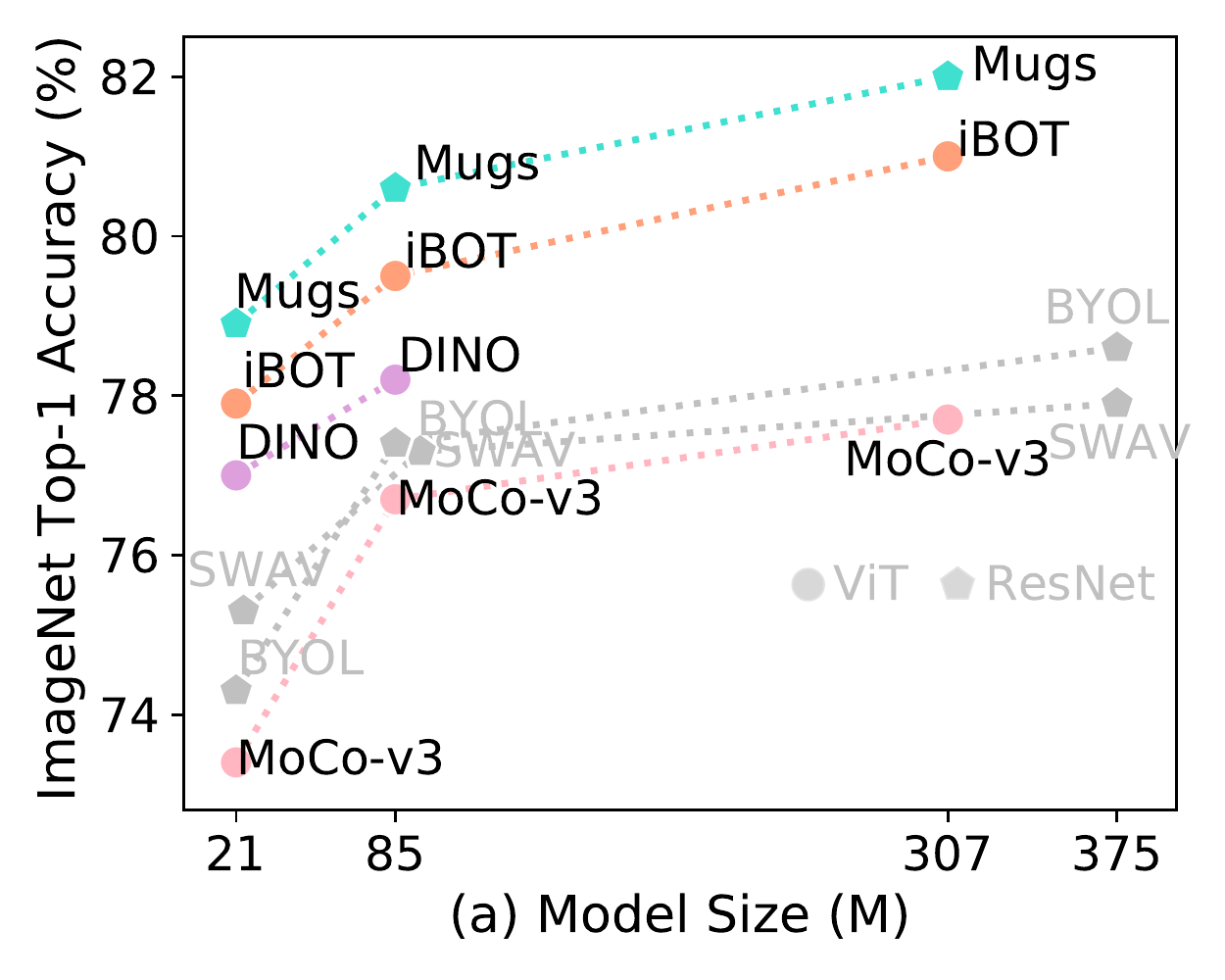}
		\hspace{-37.5mm}\resizebox{.51\columnwidth}{!}{\tablestyle{1.5pt}{1}
			\begin{tabular}[b]{llcc}
				\shline
				&\vline ViT-S/16 & ViT-B/16 & ViT-L/16  \\
				\shline
				iBOT &  \vline $  $ 77.9\%   & 79.5\% & 81.0\%  \\
				\bf Mugs &\vline $  $ \bf 78.9\% & \bf 80.6\% & \bf 82.1\% \\ 
				\shline
				\\ \\ \\ \\  
		\end{tabular}}
	\end{minipage}
	\hspace{-3.4em}
	\begin{minipage}[c]{.55\linewidth}
		\vspace{-0.45em}
		\centering
		\includegraphics[width=0.81\textwidth]{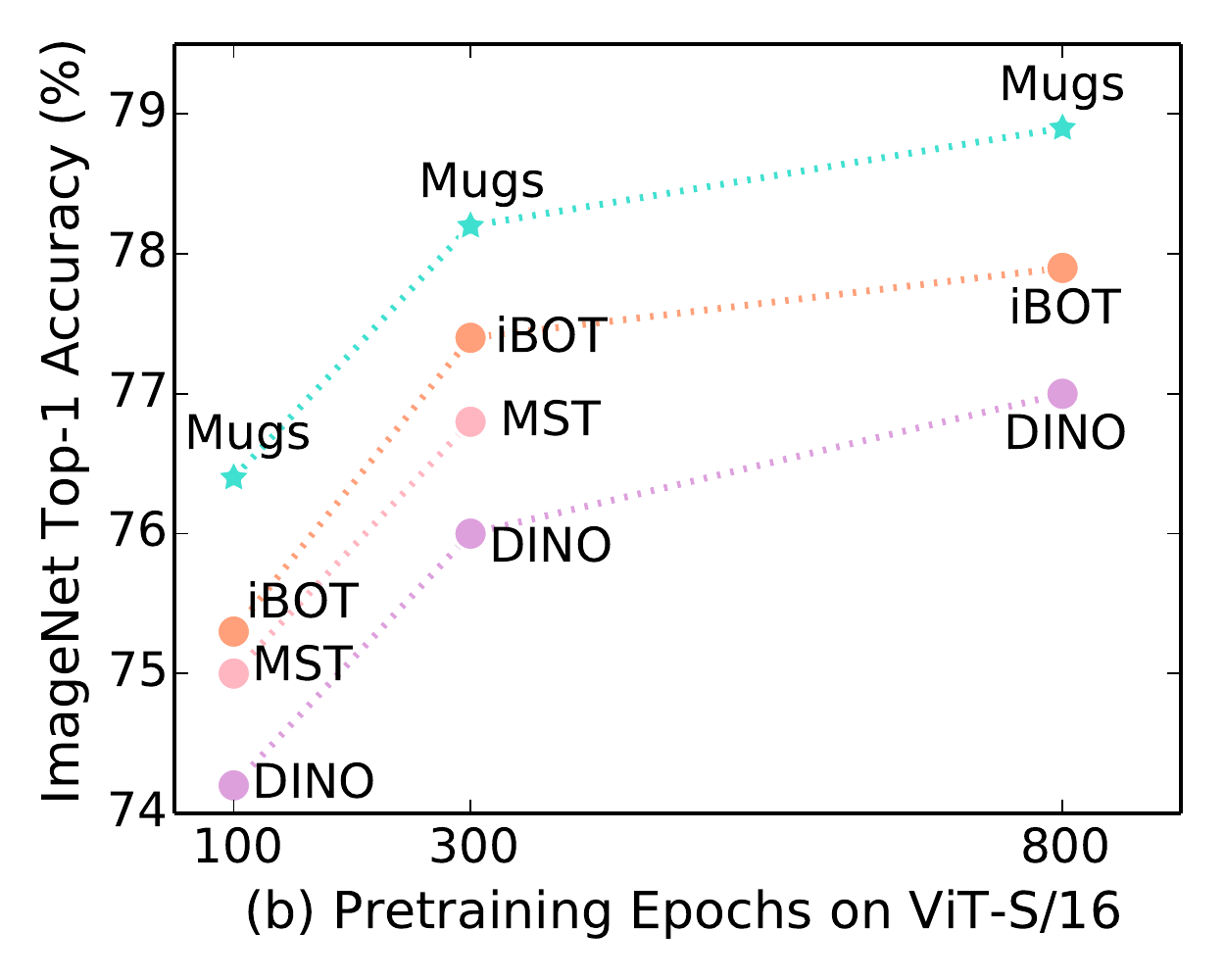}
		\hspace{-32.4mm}\resizebox{.435\columnwidth}{!}{\tablestyle{1.5pt}{1}
			\begin{tabular}[b]{llcc}
				\shline
				epochs &\vline $  $ $ $ 100  & 300 & 800  \\
				\shline
				iBOT &  \vline $  $ 75.3\%   & 77.4\% & 77.9\%  \\ 
				\bf Mugs &\vline $  $ \bf 76.4\% & \bf 78.2\% & \bf 78.9\% \\  
				\shline
				\\ \\    
				\vspace{0.5em}
		\end{tabular}}
	\end{minipage}
	\vspace{-0.5cm}
	\caption{Comparison of \textbf{linear probing} accuracy on ImageNet-1K. By pretraining  on ImageNet-1K, under different model sizes (see (a)) and pretraining epochs (see (b)),   Mugs consistently improves previous SoTA (iBOT) by at least 0.8\%. 	
	}
	\label{illustration_performance}
	\vspace{-0.1cm}
\end{figure}

As shown in Fig.~\ref{illustration_performance} (a), by only pretraining on ImageNet-1K, our Mugs sets a new state-of-the-art (SoTA)  \pz{82.1$\%$}  linear probing accuracy on  ImageNet-1K and  surpasses the previous  SoTA, \ie~iBOT~\cite{iBOT},  by a large margin \pz{$1.1\%$}.  
Moreover, under different model sizes (see Fig.~\ref{illustration_performance} (a)) and pretraining epochs (see Fig.~\ref{illustration_performance} (b)),   Mugs consistently improves previous SoTA pretrained on ImageNet 1K  
by a non-trivial \pz{0.8\%} linear probing accuracy. 
Under other three classification settings, \ie~KNN, fine-tuning the whole network and semi-supervised learning,  Mugs also beats previous best methods on the same model. 
In addition, Mugs shows its advantages on transfer learning, object detection, instance and semantic segmentation tasks. 
These results well testify the high quality, generality and transferability of the learnt features by Mugs.
Note that in this work,  we evaluate the effectiveness of Mugs through vision transformer (ViT)~\cite{ViT,Swin}, as ViT often achieves better performance than CNN of the same model size~\cite{DeiT,Swin} and also shows great potential to unify vision and language models~\cite{MAE,Alexei2022datavec}.

\vspace{-0.9em}
\section{Related works}\label{relatedwork}
\vspace{-0.7em}
As an effective family of SSL, contrastive learning~\cite{chen2020improved,hadsell2006dimensionality,hjelm2018learning,oord2018representation,bachman2019learning}, \eg, MoCo~\cite{he2020momentum} \\ and SimCLR~\cite{chen2020simple}, has gained much attention recently. 
Its key is an instance discrimination task which aims to train a network so that the positive pair, \ie~crops of the same image, are close but far from the negatives, namely the crops of other images. 
Later, BYOL \cite{BYOL} trains a network by only bringing two positives close in the feature space without using any negatives.  
Though successful, these methods only distinguish individual instances to learn fine-grained feature, and often cannot well push semantically similar instances close, impairing their performance.

Another line of SSL is clustering learning~\cite{zhou2014learning,bautista2016cliquecnn,xie2017aggregated,caron2018deep,wu2018unsupervised,huang2019unsupervised,yan2020clusterfit,lin2021prototypical}, which\\ assigns pseudo cluster labels for each sample and then trains a network to learn unsupervised representation. 
For instance, deepcluster~\cite{DeepCluster} uses k-means to cluster all data features and generates pseudo clustering labels which are then used to train a network.  
PCL~\cite{li2020prototypical} and SwAV~\cite{caron2020unsupervised}  integrate  scalable online clustering approach with contrastive learning via learning cluster prototypes as the positive and negative instances for queries, and achieve promising  performance.  
Recently, DINO~\cite{DINO} and MST~\cite{li2021mst}  propose a much simple online labeling framework that generates pseudo-labels via a  momentum teacher.  Unfortunately, this clustering family often learns cluster-level coarse-grained (semantic) features, and cannot well handle the downstream tasks which desire fine-grained features.

Finally, the recently proposed masked auto-encoder (MAE)~\cite{he2021masked,xie2021simmim,dong2021peco,dalal2005histograms,Alexei2022datavec} is a new SSL family. 
It builds on a reconstruction task that randomly masks image patches and then reconstructs the missing pixels or semantic features  
via an auto-encoder.  
However,  it emphasizes local region reconstruction, and lacks semantic discrimination ability.  
As a result,  for adapting to downstream tasks via only fine-tuning a task head at the top of the pretrained backbone, 
it  performs much worse than contrastive and clustering learning~\cite{MAE,iBOT}.  Indeed, to achieve satisfactory performance,   this reconstruction family needs to fine-tune the whole pretrained network for learning global semantics which are necessary for many downstream tasks, \eg~classification. But this requires much higher extra training cost, and also results in  very  different models for different downstream tasks, which  destroys model compatibility and increases   practical deployment cost.

\begin{figure}[t]
	\centering
	\includegraphics[width=0.99\textwidth]{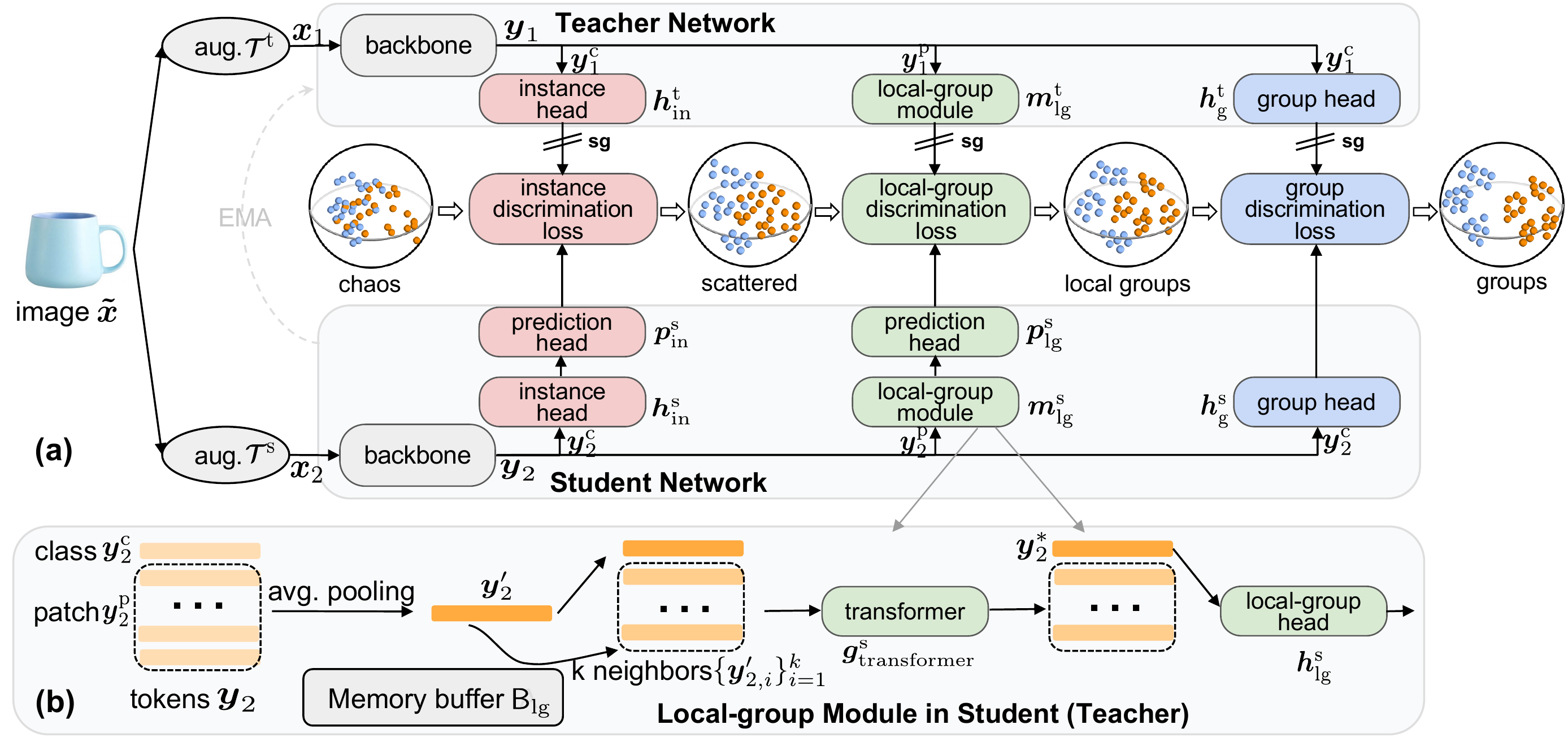}
	\vspace{-0.4em}
	\caption{\textbf{Overall framework of Mugs}. {\bf{(a)}} shows the overall framework.  For each image, Mugs respectively performs two random augmentations and feeds two crops into backbones of student and teacher. Next, it adopts three granular supervisions/losses: 1) instance discrimination supervision, 
		2) local-group discrimination supervision,  
		and 3)  group discrimination supervision. 
		Teacher is updated via exponential moving average of student.  ``sg" denotes  stop-gradient.   {\bf{(b)}} shows the pipeline of local-group modules in both student and teacher. This module averages all patch tokens, and then finds top-$k$ neighbors from memory buffer. Next, it uses a transformer to aggregate the average and its $k$ neighbors to obtain a local-group feature (class token) and feeds it into a local-group head. }
	\label{fig:MGL}
	\vspace{0.2em}
\end{figure}

\vspace{-0.8em}
\section{Multi-granular  self-supervised  learning}
\vspace{-0.3em}
Here we first introduce the overall framework of our MUlti-Granular Self-supervised learning (Mugs), and then elaborate on its three granular supervisions. 
We evaluate the effectiveness of Mugs through ViT~\cite{ViT,Swin} and thus will take ViT as an example to introduce Mugs. 
This is because 1) with similar model size, ViT shows better performance than CNN~\cite{DeiT,Swin,MAE,yu2021metaformer}; 2) ViT shows great potential for unifying the vision and language models~\cite{MAE,Alexei2022datavec}.

\vspace{-0.8em}
\subsection{Overall Framework}\label{framework}
\vspace{-0.3em}
Here we first introduce our motivation.  
As discussed in Sec.~\ref{introduction},  different downstream tasks, \eg~general  classification and its fine-grained variant,  often require different granular features, \eg~coarse- or fine-grained one. More importantly,  many real downstream tasks  actually highly desire multi-granular feature.  
For instance,  for classification problems, 
one needs  coarse-grained feature to discriminate animals from other big categories, \eg~plants, since one big category often share much coarse-grained features, \eg~legs and head in animals,  leaves and  multi-branches in plants.  Then to distinguish subcategories in each big category, \eg~dogs and wolves, a slightly lower-level coarse-grained (high-level fine-grained) feature is necessary. Finally, to further discriminate different species in each subcategory, \eg~Labrador dog and  poodle dog, more fine-grained feature is actually required.  Unfortunately, this important requirement for multi-granular features is seldom brought to front and even totally ignored in existing SSL methods.  

To alleviate this issue, we propose a simple but effective Mugs framework to learn multi-granular features, which can better satisfy different granular feature requirements of various downstream tasks and  thus enjoy higher transferability and generality than single-granular features. 
As illustrated in Fig.~\ref{fig:MGL} (a),  given an image $\xmti{}$, Mugs independently performers  augmentations $\Tt$ and $\Ts$ to obtain its two crops $\xmi{1}$ and $\xmi{2}$. 
Next, it respectively feeds $\xmi{1}$ and $\xmi{2}$ into the teacher and student backbones, and obtains their corresponding representations $\ymi{1}$ and $\ymi{2}$ which contain class and patch tokens.  Finally, Mugs builds three granular supervisions: 1) instance discrimination supervision for instance-level  fine-grained features, 2) local-group discrimination supervision for high-level fine-grained features at a local-group level, 3) group discrimination supervision for coarse-grained semantic features at a (semantic) group level.  
In this way, Mugs can learn multi-granular features, and thus better handles as many downstream tasks as possible, in contrast with SSL methods that only consider single-granular features, \eg~MoCo~\cite{MoCo-v1,MoCo-v3} for instance discriminative fine-grained features and deepclustering~\cite{DeepCluster}/DINO~\cite{DINO}~for group-discriminative coarse-grained features. Next, we will introduce our three granular supervisions which help Mugs learn multi-granular features in a complementary manner. 

\vspace{-0.7em}
\subsection{Multi-granular supervisions} 
\vspace{-0.3em}

\subsubsection{Instance discrimination supervision.}  
With this supervision,  Mugs regards each instance  as a unique class of its own which is our finest level of granularity. Accordingly, it pulls the random crops of the same instance together and pushes the crops from different images away in the feature space. In this way, it approximately scatters the instance features separately from the chaos distribution on the spherical surface as shown in Fig.~\ref{fig:MGL} (a), which also accords with the empirical and theoretical observations in~\cite{lin2021prototypical,tian2020makes} and our experimental results in Fig.~\ref{illustration_clustering} of Sec.~\ref{exp}. To implement this supervision  on the instance $\xmti{}$, Mugs respectively feeds two class tokens $\ymi{1}^{\text{c}}$ and $\ymi{2}^{\text{c}}$  in the two features $\ymi{1}$ and $\ymi{2}$ into their corresponding 
instance  heads $\hint$ and $\hins$ as shown in Fig.~\ref{fig:MGL} (a).  
Next,  
Mugs additionally passes $\hins(\ymi{2}^{\text{c}})$ into an extra prediction head $\pin$ which could alleviate the side effects of feature alignment upon the generality of the feature learnt by student or teacher backbone.  
Finally, following MoCo~\cite{MoCo-v1,MoCo-v2,MoCo-v3}, Mugs employs  InfoNCE loss~\cite{InfoNCE} for instance  discrimination:
\begin{equation}
	\setlength{\abovedisplayskip}{4pt}
	\setlength{\belowdisplayskip}{4pt}
	\setlength{\abovedisplayshortskip}{4pt}
	\setlength{\belowdisplayshortskip}{4pt}
	\LL_{\text{instance}}(\xmi{1}, \xmi{2}) \!=\! -\log\frac{\exp(\cos(\zmi{1}, \zmi{2})/\tauin)}{\exp(\cos(\zmi{1}, \zmi{2})/ \tauin) + \sum_{\zmi{}\in\Bin}\!\!\exp(\cos(\zmi{2}, \zmi{})/ \tauin)},
\end{equation}
where $\zmi{1} = \hint(\ymi{1}^{\text{c}})$, $\zmi{2} =\pin(\hins(\ymi{2}^{\text{c}}))$,  and $\tauin$ is a temperature. The buffer $\Bin$  stores the negative instances of  $\zmi{2}$, and is updated  by the historical minibatch features $\{\zmi{1}\}$ generated by teacher in a first-in and first-out order.  In this way, Mugs  pushes the crop  $\xmi{2}$ away from its negative keys (other instances) in the buffer $\Bin$ while pulling together its positive crop  $\xmi{1}$.  So it can encourage the model to learn   fine-grained features, and boost instance-level feature diversity.

\vspace{-1.3em}
\subsubsection{Local-group discrimination supervision.}   As explained in Sec.~\ref{framework}, fine-grained features are often insufficient for diverse  downstream tasks, \eg~classification, due to lack of sufficient high-level data semantics.  
In this work, to learn higher-level fine-grained features, also called ``local-group features" here, Mugs proposes a novel and effective local-group supervision. Intuitively, as shown in the third sphere of Fig.~\ref{fig:MGL} (a), this supervision encourages instance features to have small but separately scattered local-group structures, \ie~small/large distance among highly similar/dissimilar samples. 
See detailed explanation at the end of this subsection.  Accordingly, it  helps capture slightly higher-level semantics in the data compared with instance discrimination supervision. Next, we  will first introduce its details and then discuss its benefits to Mugs. 

As shown in  Fig.~\ref{fig:MGL} (a), given the crop $\xmi{1}$ of image $\xmti{}$,  the teacher backbone outputs $\ymi{1}$ which contains class token $\ymi{1}^{\text{c}}$ and patch tokens $\ymi{1}^{\text{p}}$. Similarly, Mugs feeds another crop  $\xmi{2}$ of $\xmti{}$ into student backbone to obtain  $\ymi{2}$ consisting of class token $\ymi{2}^{\text{c}}$ and patch tokens $\ymi{2}^{\text{p}}$. Next, Mugs respectively averages the patch tokens $\ymi{1}^{\text{p}}$ and $\ymi{2}^{\text{p}}$ to obtain two average tokens $\ymsi{1}$ and $\ymsi{2}$ as illustrated in Fig.~\ref{fig:MGL} (b). Here we use the average of patch tokens instead of the class token, as we find that the average token often contains more information and finds more accurate neighbors in the next step.  
Note, for a CNN backbone, we can average the last feature map to obtain $\ymsi{1}$ and $\ymsi{2}$.  On the other hand, Mugs uses a memory buffer $\Blg$ to store the historical minibatch average tokens $\{\ymsi{1}\}$ and $\{\ymsi{2}\}$ in a first-in and first-out order. Then as shown by Fig.~\ref{fig:MGL} (b),  for both $\ymsi{1}$ and $\ymsi{2}$, Mugs  respectively finds their corresponding top-$k$  neighbors $\{\ymsi{1, i}\}_{i=1}^k$ and $\{\ymsi{2, i}\}_{i=1}^k$ from the buffer $\Blg$.  Next, it respectively uses a transformer to aggregate the average token and its $k$  neighbors as follows: 
\begin{equation}
	\setlength{\abovedisplayskip}{4pt}
	\setlength{\belowdisplayskip}{4pt}
	\setlength{\abovedisplayshortskip}{4pt}
	\setlength{\belowdisplayshortskip}{4pt}
	\ymfi{1} = \gt(\ymsi{1}; \{\ymsi{1, i}\}_{i=1}^k)  \qquad \text{and} \qquad \ymfi{2} = \gs(\ymsi{2}; \{\ymsi{2, i}\}_{i=1}^k).
\end{equation}
Here $\gt(\ymsi{1}; \{\ymsi{1, i}\}_{i=1}^k)$ denotes a $2$-layered vanilla ViT without any patch  embedding layers,
and has input class token $\ymsi{1}$, input patch tokens $\{\ymsi{1, i}\}_{i=1}^k$ and output class token $\ymfi{1}$. Since the new representation $\ymfi{1}$ comes from $\ymsi{1}$ and its top-$k$  neighbors $\{\ymsi{1, i}\}_{i=1}^k$ which together constitute a local group of $\ymsi{1}$, $\ymfi{1}$ is also called a ``local group feature". $\gs(\ymsi{2}; \{\ymsi{2, i}\}_{i=1}^k)$ has the same function.
Finally, Mugs pulls these two local-group features $\ymfi{1}$ and $\ymfi{2}$ from the same instance $\xmti{}$ together and pushes away the local-group features of other instances.  Similarly, Mugs adopts the InfoNCE loss~\cite{InfoNCE} to achieve this target: 
\begin{equation}
	\setlength{\abovedisplayskip}{4pt}
	\setlength{\belowdisplayskip}{4pt}
	\setlength{\abovedisplayshortskip}{4pt}
	\setlength{\belowdisplayshortskip}{4pt}
	\LL_{\text{local-group}}(\xmi{1}, \xmi{2}) \!=\! -\log\frac{\exp(\cos(\zmi{1}, \zmi{2})/ \taulg)}{\exp(\cos(\zmi{1}, \zmi{2})/ \taulg) \!+\! \sum_{\zmi{}\in\Blg'}\!\!\!\exp(\cos(\zmi{2}, \zmi{})/ \taulg)},
\end{equation}
where $\zmi{1} = \hlgt(\ymfi{1})$ and $\zmi{2} =\plg(\hlgs(\ymfi{2}))$.  $\hlgt$ and $\hlgs$ are two projection heads and $\plg$ is a prediction head. 
Buffer $\Blg'$  also stores the historical local-group features $\{\ymfi{1}\}$  produced by teacher in a first-in first-out order.  

This supervision benefits Mugs from two aspects.  \textit{Firstly}, it provides complementary instance supervision to the above instance discrimination supervision. It brings two local-group features $\ymfi{1}$ and $\ymfi{2}$ from the same instance $\xmt{}$ together, where local-group features $\ymfi{1}$/$\ymfi{2}$ are the aggregation of the crop $\xmi{1}$/$\xmi{2}$ and its top-k nearest neighbors. So to achieve small local-group supervision loss $	\LL_{\text{local-group}}(\xmi{1}, \xmi{2})$, the two crops $\xmi{1}$ and $\xmi{2}$ of $\xmti{}$ should have very similar top-k neighbors. This means besides the crops themselves, their corresponding neighbors should also be well aligned, which is an extra challenging alignment problem compared with instance discrimination supervision and enhances local-group semantic alignment. 
\textit{Secondly}, it encourages highly-similar instances to form local-groups and scatters these local-groups separately, increasing the semantic discrimination ability of the learnt feature.   
This is because 1) this supervision uses a small $k$ (around 10) for neighbors such that the samples in the same local-group are highly similar and have small distance, helping form local-groups; 2)   
this supervision further pushes current local-group feature away local-group features from other instances, and thus scatters different local-groups separately.  
With these two aspects, this local-group discrimination supervision boosts higher-level fine-grained feature learning by considering the local-group structures in data. 

\vspace{-1.3em}
\subsubsection{Group discrimination supervision.}  This supervision is the most coarse level supervision in Mugs. Intuitively, as shown in the last sphere in Fig.~\ref{fig:MGL} (a), it targets at clustering semantically similar instances and  local-groups into the same big group/cluster which could reveal more \textit{global} semantics in data compared with the instance and local-group supervisions. 
In the following, we first introduce its details and then discuss the co-effects of the three granular supervisions. 

For the instance $\xmti{}$, Mugs respectively feeds  the class token $\ymi{1}^{\text{c}}$ in the feature $\ymi{1}$ from teacher backbone and the class token $\ymi{2}^{\text{c}}$  in $\ymi{2}$ from student backbone into two group heads $\hsgt$ and $\hsgs$.  Then, it builds a set of learnable cluster prototypes $\{\cmi{i}\}_{i=1}^{m}$ and computes  soft pseudo clustering labels in an online manner:
\begin{equation}\label{pseudolabel}
	\setlength{\abovedisplayskip}{3pt}
	\setlength{\belowdisplayskip}{3pt}
	\setlength{\abovedisplayshortskip}{3pt}
	\setlength{\belowdisplayshortskip}{3pt}
	\pt_i=\frac{\exp(\sigma(\hsgt(\ymi{1}^{\text{c}})) \cdot \cmi{i}/\tausg)}{\sum_{i=1}^{m} \exp(\sigma(\hsgt(\ymi{1}^{\text{c}})) \cdot \cmi{i}/\tausg)}      \ \ \text{and} \ \  \ps_i=\frac{\exp(\hsgs(\ymi{2}^{\text{c}}) \cdot \cmi{i}/\tausg')}{\sum_{i=1}^{m} \exp(\hsgs(\ymi{2}^{\text{c}}) \cdot \cmi{i}/\tausg')}. 
\end{equation}
Here the function $\sigma(\hsgt(\ymi{1}^{\text{c}})) = \hsgt(\ymi{1}^{\text{c}})- \ptm$ is to increase the diversity of the feature $\hsgt(\ymi{1}^{\text{c}})$ and thus sharpens the soft pseudo label $\pt$ in Eqn.~\eqref{pseudolabel}, where $\ptm$ denotes the estimated average statistics of all past $\hsgt(\ymi{1}^{\text{c}})$ via an exponential moving average of the mini-batch $\B$, namely, $\ptm \leftarrow \rho \cdot \ptm  + (1-\rho)\cdot \frac{1}{|\B|}\sum_{\pt\in\B} \pt$ with a constant $\rho \in [0, 1]$. Such a technique is shown to be useful in~\cite{DINO}.   
Next, similar to a supervised classification task, Mugs employs the cross-entropy loss but with soft labels as its training loss:
\begin{equation}
\setlength{\abovedisplayskip}{3pt}
\setlength{\belowdisplayskip}{3pt}
\setlength{\abovedisplayshortskip}{3pt}
\setlength{\belowdisplayshortskip}{3pt}
	\LL_{\text{group}}(\xmi{1}, \xmi{2}) = -\sum\nolimits_{i=1}^{m} \pt_i \log \left(\ps_i\right).
\end{equation}
Now we put our three granular supervisions together and discuss their co-effects on representation learning which also distinguishes it from existing methods~\cite{MoCo-v3,DINO}. As aforementioned, instance discrimination supervision is to pull the crops of the same image together and to approximately scatter the instance features separately on the spherical surface as shown in Fig.~\ref{fig:MGL} (a). It helps Mugs learn instance-level fine-grained features.  Next, the local-group discrimination supervision first provides complementary supervision for instance discrimination supervision by encouraging crops of the same instance to have highly similar neighbors. 
Then, as shown in the third sphere in Fig.~\ref{fig:MGL} (a), the local-group supervision scatters different local-groups formed by crops and its neighbors separately to boost the semantic discrimination ability of these local-groups. This supervision mainly learns higher-level fine-grained  features at a local-group level. Finally,  to avoid similar local-groups to be scattered randomly or far away, the group discrimination supervision brings similar samples together and thus pulls similar local-groups close, as intuitively illustrated by the last sphere in Fig.~\ref{fig:MGL} (a).  It is responsible to capture the coarse-grained features at a (semantic) group level.  With these three granular supervisions, Mugs can well learn three different but complementary granular features, which are characterized by better generality and transferability on the various kinds of downstream tasks compared with single-granular features. Compared with existing methods, \eg~MoCo~\cite{MoCo-v3} and DINO~\cite{DINO}, the main novelties  of Mugs   lie in two folds: 1) Mugs   learns  multi-granular representation via three complementary supervisions and  can often better handle  diverse downstream tasks than the  existing methods that often learn single-granular feature; 2) Mugs proposes a novel  and effective local-group supervision which complements both instance and group discrimination supervisions and benefits Mugs from two aspects as discussed above.

\vspace{-1.2em}
\subsubsection{Overall training objective.}  Now we introduce the overall training loss:
\begin{equation}\label{overallloss}
	\setlength{\abovedisplayskip}{5pt}
	\setlength{\belowdisplayskip}{5pt}
	\setlength{\abovedisplayshortskip}{5pt}
	\setlength{\belowdisplayshortskip}{5pt}
	\LL(\xmi{1}, \xmi{2}) \!=\!  \lamin \LL_{\text{instance}}(\xmi{1}, \xmi{2}) 
	+ \lamlg    \LL_{\text{local-group}}(\xmi{1}, \xmi{2}) + \lamsg  \LL_{\text{group}}(\xmi{1}, \xmi{2}),
\end{equation}
where the three constants $\lamin$, $\lamlg$ and $\lamsg$ are to trade-off the three supervisions. For simplicity, we set  $\lamin\!=\!\lamlg\!=\!\lamsg\!=\!\frac{1}{3}$ in all experiments. We then can minimize the objective 	$\LL(\xmi{1}, \xmi{2})$  to optimize  student network.   Teacher network  is updated   via the exponential moving average  of corresponding  parameters in student.

\vspace{-0.8em}
\section{Experiments}\label{exp}
\vspace{-0.4em}
Here we present the performance evaluation of our Mugs on benchmark classification tasks, 
transfer learning task,  delectation and  segmentation  tasks, and video segmentation task with comparison against several representative state-of-the-art SSL approaches.  
\textit{The code and models will be released soon. } 

\vspace{-1.2em}
\subsubsection{Architectures.}   
As aforementioned, we  use ViT~\cite{ViT}  to evaluate our Mugs.   
For the instance and local-group discriminations, their four projection heads, \ie, $\hint$, $\hins$, $\hlgt$ and $\hlgs$,  are all 3-layered MLPs with hidden/output dimension 2,048/256, and their prediction heads $\pin$ and $\plg$  are all 2-layered MLPs with hidden/output dimension 4,096/256.  For group discrimination, its projection heads $\hsgt$ and $\hsgs$, are 3-layered MLP with hidden/output dimension of 2,048/256. Transformers $\gt$ and $\gs$ have 2 layers and have a total input token number of 9  as we set $k=8$ for the neighbors.  For the three buffers ($\Bin$,   $\Blg$ and $\Blg'$) and the prototypes $\{\cmi{i}\}_{i=1}^{m}$,  their sizes are all 65,536.

\vspace{-1.2em}
\subsubsection{Pretraining setup.} We pretrain Mugs on the training data of ImageNet-1K~\cite{ImageNet} without labels.  Following~\cite{iBOT}, we pretrain ViT-S/16 for 800 epochs, ViT-B/16 for 400 epochs,  and ViT-L for 250 epochs.  Following DINO and iBOT, 
we  use symmetric training loss, \ie~$\frac{1}{2}(\LL(\xmi{1}, \xmi{2}) + \LL(\xmi{2}, \xmi{1}))$.  We use AdamW optimizer~\cite{AdamW}  with  a momentum of 0.9,  a weight decay of 0.1, and  a cosine schedule~\cite{SGDR}. 
For data augmentation, we adopt weak augmentation in DINO~\cite{DINO}  to implement \scalebox{0.85}{$\Tt$} in teacher, and use strong augmentation (mainly including AutoAugment~\cite{AutoAugment}) in 
DeiT~\cite{DeiT} as the augmentation  \scalebox{0.85}{$\Ts$} in student.  
Following conventional multi-crop setting~\cite{SwAV,DINO,iBOT}, we crop each image into 2 large crops of size $224$ and 10 extra small crops of size $96$. For both large crops, we feed each of them into teacher and use its output to supervise the student’s output from the other 11 crops.  For two-crop setting, Table~\ref{tab:nomulticrop} in Appendix~\ref{moreResults} reports the results and shows the superiority of Mugs over SoTAs. 

For Mugs, we follow MoCo to set $\tauin=\taulg=0.2$ in the infoNCE loss, and follow DINO to set $\tausg'=0.1$ and linearly warm up $\tausg$ from $0.04$ to $0.07$. We set the neighbor number $k=8$, and set  $\rho=0.9$  
in group discrimination. 
Mugs has almost the same training cost with DINO, \eg~about $27$  hours with $8$ A100  GPUs for  $100$ pretraining epochs on ViT-S/16, as our projection/prediction heads and  transformers $\gtrans$  are much smaller than the backbone.   See more details of the augmentation, multi-crop  loss, and pretraining cost in Appendix \ref{moreexp}.

\begin{table}[t]
	\begin{center}
		\caption{\textbf{Linear probing and k-NN} accuracy (\%) on ImageNet-1K. ``Dataset" denotes which dataset is used to pretrain. ``Epo." is the effective pretraining epochs adjusted by number of views processed by the models following~\cite{iBOT}.}
		\label{tablelinear}
		\setlength{\tabcolsep}{5.6pt} 
		\renewcommand{\arraystretch}{1.0}
		{ \fontsize{8.3}{3}\selectfont{
				\begin{tabular}{l c c cc c c}
					\toprule
					\textbf{Method}  &  \textbf{Arch.}   & \textbf{\#Params} & \textbf{Dataset} & \textbf{Epo.} & \textbf{Lin.} & \textbf{k-NN}  \\
					\midrule 
					MoCo-v3~\cite{MoCo-v3}   & ResNet-50 & 23M &  ImageNet-1K  & 1600 &  74.6 & --- \\
					SimCLR~\cite{SimCLR}   & ResNet-50 & 23M &  ImageNet-1K  & 1600 &  69.3 & --- \\
					InfoMin Aug~\cite{tian2020makes}& ResNet-50 & 23M &  ImageNet-1K  & 1600 &  73.0 & --- \\
					SimSiam~\cite{chen2021exploring}         & ResNet-50 & 23M & ImageNet-1K  & 1600 & 71.3 & --- \\
					BYOL~\cite{BYOL}   & ResNet-50 & 23M &  ImageNet-1K  & 2000 &  74.3 & --- \\
					SwAV~\cite{SwAV}         & ResNet-50 & 23M & ImageNet-1K  & 2400 & 75.3 & 65.7 \\
					DeepCluster-v2~\cite{DeepCluster}   & ResNet-50 & 23M & ImageNet-1K  & 2400 & 75.2 & ---\\
					DINO~\cite{DINO}         & ResNet-50 & 23M&  ImageNet-1K   & 3200 & 75.3 & 67.5 \\
					\midrule
					MoCo-v3~\cite{MoCo-v3}  & ViT-S/16 & 21M & ImageNet-1K & 3200 & 73.4 &  --- \\
					SwAV~\cite{SwAV}        & ViT-S/16 & 21M & ImageNet-1K & 3200 & 73.5 & 66.3 \\
					DINO~\cite{DINO}        & ViT-S/16 & 21M & ImageNet-1K & 3200 & 77.0 & 74.5 \\
					iBOT~\cite{iBOT}        & ViT-S/16 & 21M & ImageNet-1K & 3200 & 77.9 & 75.2 \\
					\textbf{Mugs \scriptsize{(ours)}}     & ViT-S/16 & 21M & ImageNet-1K & 3200 & \textbf{78.9} &  \textbf{75.6} \\
					\midrule
					MoCo-v3~\cite{MoCo-v3}  & ViT-B/16 & 85M & ImageNet-1K & 1200 & 76.7 &  --- \\
					DINO~\cite{DINO}        & ViT-B/16 & 85M & ImageNet-1K & 1600 & 78.2 & 76.1 \\
					iBOT~\cite{iBOT}        & ViT-B/16 & 85M & ImageNet-1K & 1600 & 79.5 & 77.1 \\
					\textbf{Mugs \scriptsize{(ours)}}     & ViT-B/16 & 85M & ImageNet-1K & 1600 & \textbf{80.6} &  \textbf{78.0} \\        
					\midrule
					MoCo-v3~\cite{MoCo-v3}  & ViT-L/16 & 307M & ImageNet-1K & 1200 & 77.6 &  --- \\
					iBOT~\cite{iBOT}        & ViT-L/16 & 307M & ImageNet-1K & 1000 & 81.0  & 78.0 \\
					\textbf{Mugs \scriptsize{(ours)}}     & ViT-L/16 & 307M & ImageNet-1K & 1000 & \textbf{82.1} & \textbf{80.3} \\        
					\bottomrule
					iBOT~\cite{iBOT}        & ViT-L/16 & 307M & ImageNet-22K & 200  & 82.3  & 72.9  \\
					\bottomrule 
		\end{tabular}}}
	\end{center}
	\vspace{-1.4em}
\end{table}

\begin{table}[t]
	\caption{\textbf{Fine-tuning}  classification accuracy (\%)  on  ImageNet-1K. All methods are pretrained on ImageNet-1K. ``Epo." is the effective pretraining epochs adjusted by number of views processed by the models following~\cite{iBOT}.\vspace{-1.0em}}
	\label{tablefinetune}
	\begin{center}
		\setlength{\tabcolsep}{6.6pt} 
		\renewcommand{\arraystretch}{1.0}
		{ \fontsize{8.3}{3}\selectfont{
				\begin{tabular}{c | l c c| cc  }
					\toprule
					&	\multirow{2}{*}{\textbf{Method}}  &  \multicolumn{2}{c|}{\textbf{ViT-S/16}}  &  \multicolumn{2}{c}{\textbf{ViT-B/16}} \\
					&		\textbf{}  & {Epo.} & {Acc. (\%)}  & {Epo.} & {Acc. (\%)}  \\
					\midrule 
					&	Supervised~\cite{DeiT}  &  ---    & 79.9   & ---    & 81.8  \\
					\hline
					
					&  		BEiT~\cite{BEiT}     & 800  & 81.4   & 800  & 83.4    \\
					&MAE~\cite{MAE}      & ---& --- & 1600 & 83.6  \\
					
					reconstruction 	&SimMIM~\cite{xie2021simmim}       &--- &   ---  & 1600 & 83.8   \\ 
					& MaskFeat~\cite{wei2021masked}      & ---&   ---  & 1600 & 84.0    \\
					&	data2vec~\cite{Alexei2022datavec}    & ---&  ---    & 1600 & 84.2  \\
					\hline
					
					&MoCo-v3~\cite{MoCo-v3}  & 600  & 81.4 & 600  & 83.2  \\ 
					contrastive or		&	DINO~\cite{DINO}    & 3200 & 82.0  & 1600 & 83.6    \\
					clustering	& 	iBOT~\cite{iBOT}      & 3200 & 82.3   & 1600 & 83.8  \\
					&	\textbf{Mugs \scriptsize{(ours)}}    & 3200 & \textbf{82.6}   & 1600 & \textbf{84.3}     \\   
					\bottomrule
		\end{tabular}}}
	\end{center}
	\vspace{-1.8em}
\end{table}

\vspace{-0.9em}
\subsection{Results on ImageNet-1K}\label{resultsonImagenet}
\vspace{-0.4em}
Here we compare Mugs with SoTAs on ImageNet-1K under four classification settings, \ie, linear probing, KNN,  fine-tuning, and semi-supervised learning. 

\vspace{-1.5em}
\subsubsection{Linear Probing.} It   
trains a linear classifier   on top of frozen features generated by the backbone, \eg~ViT, for 100 epochs on ImageNet-1K.  
We follow 
DINO and iBOT, and use SGD with different learning rates for different models. Table~\ref{tablelinear} reports  top-1 test accuracy on ImageNet-1K. One can find that by pretraining on ImageNet-1K,  Mugs consistently outperforms other methods on different backbones of various  sizes. Specifically,  Mugs respectively achieves 78.9\% and 80.6\% top-1 accuracy on  ViT-S and ViT-B, and improves corresponding SoTAs by at least  1.0\%. Notably, on ViT-L,  by only pretraining on ImageNet-1K, Mugs sets a new SoTA  accuracy of \pz{82.1\%}, which is even comparable to  the   accuracy 82.3\%  pretrained on ImageNet-22K.  

\vspace{-1.2em}
\subsubsection{KNN.} It is another powerful evaluation protocol to test the quality of representation learnt by SSL models. Following DINO and iBOT, we also sweep over different numbers (10, 20, 50, 100) of nearest neighbors  for each model. From Table~\ref{tablelinear}, one can find that for all backbones, Mugs achieves the highest top-1 accuracy on ImageNet-1K test dataset. Particularly,  it respectively makes   0.4\%,  0.9\%, and \pz{2.3\%}  improvement on  ViT-S, ViT-B and ViT-L over the runner-up,  showing the advantages of multi-granular representation in Mugs.

\vspace{-1.2em}
\subsubsection{Fine-tuning.}  
Besides linear probing and KNN,  fine-tuning is proposed recently to further testify performance of a pretrained model. It allows for optimizing the pretrained backbone with a linear classifier.  Following BEiT~\cite{BEiT}, DINO and iBOT, we use AdamW optimizer with layer-wise learning rate decay to train ViT-S/ViT-B/ViT-L for 200/100/50 epochs on ImageNet-1K. Table~\ref{tablefinetune} summarizes the classification results, in which ``Supervised" means randomly initializing model parameters instead of using pretrained backbone and its results are quoted from DeiT~\cite{DeiT} that trains 300 epochs. 
On  ViT-S and ViT-B, Mugs  respectively achieves new SoTA results of 82.5\% and 84.3\%, improving the runner-up, namely, iBOT and data2vec~\cite{Alexei2022datavec}, by  0.2\% and 0.1\%  respectively. Note, the reconstruction  frameworks, \eg~MAE~\cite{MAE}, MaskFeat~\cite{wei2021masked} and data2vec,  have  unsatisfactory  linear probing  performance  and thus are  included   in Table~\ref{tablelinear}. Moreover, as explained at the end of Sec.~\ref{relatedwork},  this fine-tuning setting  needs much  higher extra training cost, and also destroys model compatibility for  deployment. So here we do not further push Mugs's limits on the large models.

\begin{table}[t]
	\begin{center}
		\caption{\textbf{Semi-supervised} classification accuracy (\%) on ImageNet-1K.\vspace{-0.1em}}
		\label{table:low-shot}
		\setlength{\tabcolsep}{6.6pt} 
		\renewcommand{\arraystretch}{0.85}
		{ \fontsize{8.3}{3}\selectfont{
				\begin{tabular}{l c || c c | cc}
					\toprule
					\multirow{2}{*}{\textbf{Method}}  & \multirow{2}{*}{\textbf{Arch.}} &   \multicolumn{2}{c|}{\textbf{logistic regression}} &    \multicolumn{2}{c}{\textbf{fine-tuning}} \\
					
					& & 1\% & 10\% &    1\% & 10\%  \\
					\midrule 
					SimCLRv2~\cite{SimCLRv2} & RN50 & --- & --- & 57.9 & 68.1  \\
					BYOL~\cite{BYOL} & RN50 & ---  & ---  & 53.2 & 68.8 \\
					SwAV~\cite{SwAV} & RN50 &---  & ---  & 53.9 & 70.2 \\
					SimCLRv2+SD~\cite{SimCLRv2} & RN50 & --- & --- & 60.0 & 70.5 \\
					DINO~\cite{DINO}  & ViT-S/16 &   64.5 & 72.2  &    60.3 & 74.3\\
					iBOT~\cite{iBOT}    & ViT-S/16 &   65.9 & 73.4  &    61.9 & 75.1  \\
					\textbf{Mugs \scriptsize{(ours)}}  & ViT-S/16   &   \textbf{66.9} & \textbf{74.0} &  \textbf{66.8} & \textbf{76.8}\\
					\bottomrule
				\end{tabular}
		} }
	\end{center}
	\vspace{-1.1em}
\end{table}

\vspace{-1.3em}
\subsubsection{Semi-supervised learning.} Here we respectively use 1\%/10\% training data of ImageNet-1K to fine tune the pretrained backbones, and then evaluate on the test data.  Following~DINO and iBOT,  we consider two settings: 1) training a logistic regression classifier on frozen features; and 2) fine-tuning the whole pretrained backbone.   Table~\ref{table:low-shot} shows that for both 1\% and 10\% training data,  our Mugs  consistently surpasses previous SoTAs  under both two settings. Notably, under fine-tuning setting with 1\% labeled data,  Mugs improves iBOT by a significant 4.9\% accuracy, and shows its advantages  for low-shot learning.

\vspace{-1.3em}
\subsubsection{Result Analysis. }  Fig.~\ref{illustration_clustering} uses T-SNE~\cite{van2008visualizing} to reveal the feature differences among MoCo-v3, DINO, iBOT, and Mugs, in which each color denotes a unique class. From the last subfigure, one can find that for one class, Mugs often divides it into several small clusters in the feature space, such as 6 clusters for brown, 4  for purple,  6  for red, and 5 for blue, and scatters these small clusters in a big class. This partially reveals multi-granular structures in the feature:   classes are separately scattered, which corresponds to a group-level coarse granularity; several small  scattered clusters in a class show a local-group-level fine granularity; and some separate instances in a cluster reveal an instance-level fine granularity. In contrast, MoCo-v3, DINO and iBOT often do not show this multi-granular feature structure.  
Hence, for some challenging classes denoted by green, red and purple colors, our Mugs can well distinguish them, while  MoCo-v3, DINO and iBOT cannot. This is because instead of regarding the class as a whole, Mugs utilizes its multi-granular supervisions to consider the multi-granular (hierarchical)  data semantic structures and divide the whole class into several easily-distinguishable clusters  in the pretraining phase. Differently, MoCo-v3, DINO and iBOT ignore the multi-granular semantic  structures and only uses one granular supervision which often could not well handle the challenging classes.  Fig.~\ref{illustration_attention} (a) further visualizes the self-attention  of ViT-B/16. One can observe  Mugs can well  capture  object shapes and thus their semantics.  See more details and examples in Appendix~\ref{attention_Visualization}. 

\begin{figure*}[tb]
	\begin{center}
		\setlength{\tabcolsep}{0.8pt}   
		\begin{tabular}{cccc}
			\includegraphics[width=0.23\linewidth]{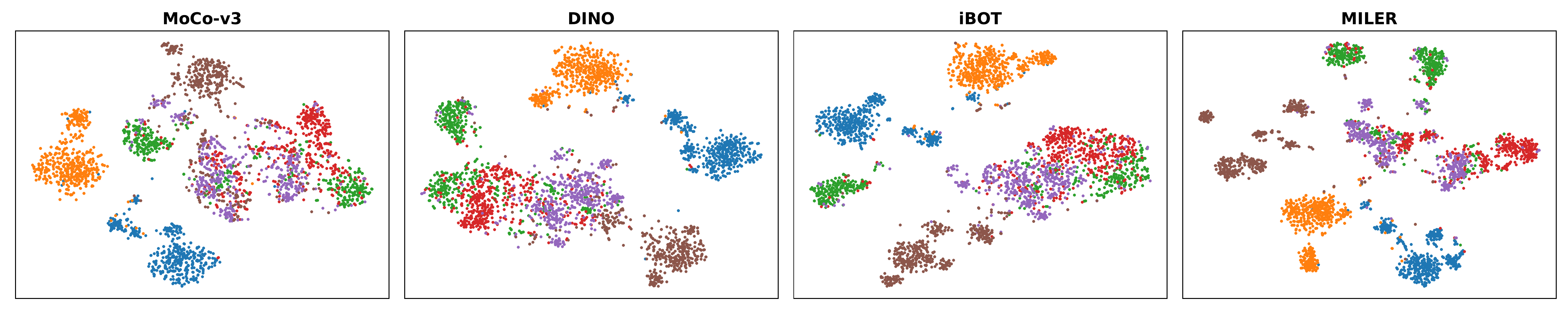} &  \includegraphics[width=0.23\linewidth]{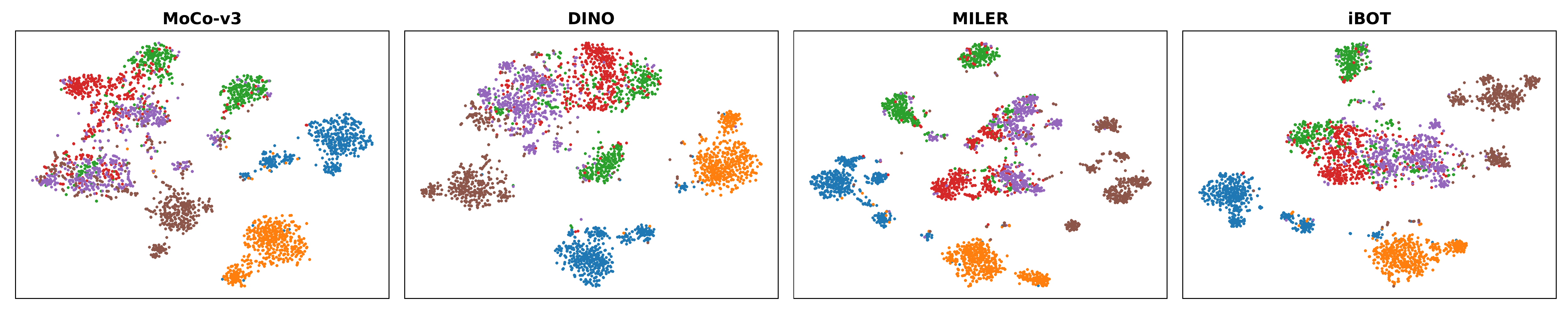} &
			\includegraphics[width=0.23\linewidth]{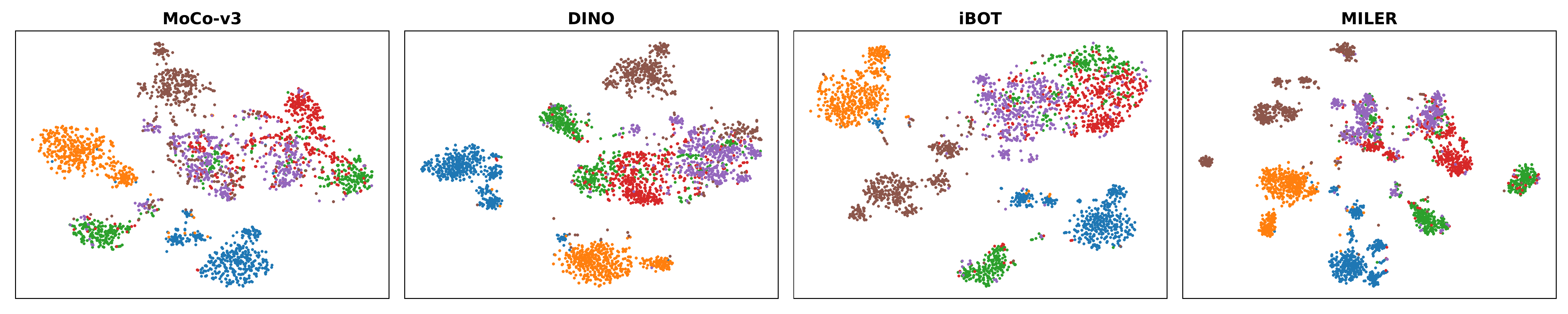} &
			\includegraphics[width=0.23\linewidth]{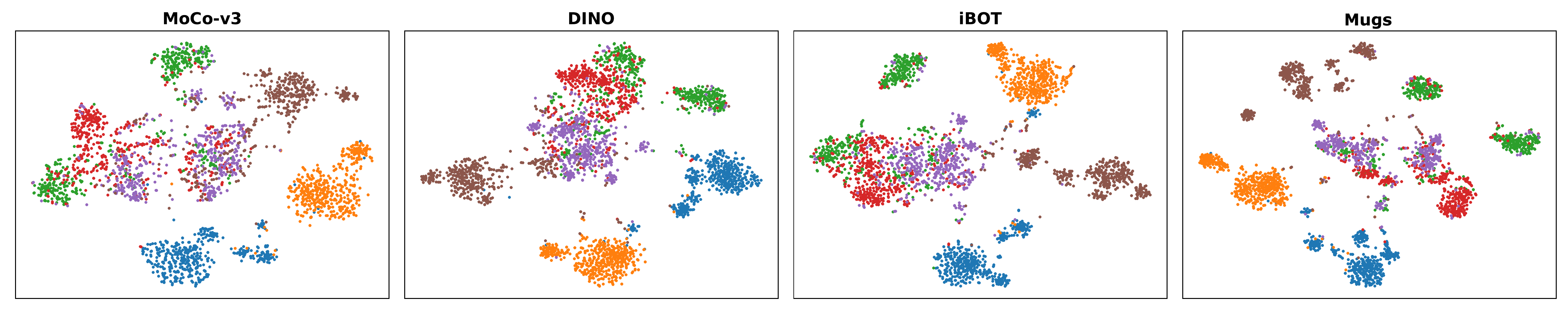} \vspace{-0.3em} 
		\end{tabular} 
	\end{center}
	\vspace{-1.6em}
	\caption{T-SNE visualization of the  learned feature  by ViT-B/16. We show  the fish classes  in ImageNet-1K, \ie, the first  six classes,  including tench, goldfish, white shark, tiger shark, hammerhead, electric ray. See more examples in Appendix. }
	\label{illustration_clustering}
	\vspace{-1.1em}
\end{figure*}

\vspace{-1.1em}
\subsection{Results on downstream tasks}
\vspace{-0.5em}
Due to limited space, we defer  training and implementation details to Appendix.  

\begin{table}[t]
	\begin{center}
		\caption{Classification accuracy (\%) for \textbf{transfer learning}  on six datasets. }
		\label{table:transfer}
		\setlength{\tabcolsep}{1.1pt} 
		\renewcommand{\arraystretch}{1.2}
		{ \fontsize{8.2}{3}\selectfont{
				\begin{tabular}{l || cccccc || cccccc}
					\toprule
					\multirow{2}{*}{\textbf{Method}} &  \multicolumn{6}{c||}{\textbf{ViT-S/16}} &  \multicolumn{6}{c}{\textbf{ViT-B/16}} \\ 
					& {{Cif$_{10}$}} & {{Cif$_{100}$}} & {{INat$_{18}$}} & {{INat$_{19}$}} & {{Flwrs}} & {{Car}} &  {{Cif$_{10}$}} & {{Cif$_{100}$}} & {{INat$_{18}$}} & {{INat$_{19}$}} & {{Flwrs}} & {{Car}} \\ 
					\hline
					Sup.~\cite{DINO}      & 99.0 & 89.5 & 70.7 & 76.6 & 98.2 & 92.1  &  {99.0} & 90.8 & {73.2} & {77.7} &{98.4} & {92.1} \\ 
					BEiT~\cite{BEiT}       & 98.6 & 87.4 & 68.5 & 76.5 & 96.4 & 92.1 & 99.0 & 90.1 & 72.3 & 79.2 & 98.0 & 94.2 \\
					MAE~\cite{MAE}       & --- & --- & --- & --- & --- & --- & --- & --- & 75.4 & 80.5  & --- & --- \\
					MoCo-v3~\cite{MoCo-v3} & --- & --- & --- & --- & --- & --- & 98.9 & 90.5 & ---     & ---    & 97.7 & ---     \\
					DINO~\cite{DINO}       & 99.0 & 90.5 & 72.0 & 78.2 & 98.5 & 93.0  & 99.1 & 91.7 & 72.6 & 78.6 & 98.8 & 93.0 \\
					iBOT~\cite{iBOT}       & 99.1 & 90.7 & 73.7 & 78.5 & 98.6 & {\bf 94.0}  &  99.2 & 92.2 &  74.6 &  79.6 &  {\bf 98.9} &  {\bf 94.3} \\
					\textbf{Mugs \scriptsize{(ours)}}    & {\bf 99.2} &  {\bf91.8} & {\bf74.4}  & {\bf 79.8} &{\bf 98.8} &  { 93.9} &  {\bf 99.3 }&	 {\bf92.8}  &  {\bf76.4} &  {\bf80.8} & {\bf98.9}  & 94.0 \\
					\bottomrule
		\end{tabular}}}
	\end{center}
	\vspace{-1.1em}
\end{table}

\vspace{-1.3em}
\subsubsection{Transfer learning.}      Here we investigate the  transferability of  the models pretrained by Mugs. Specifically,  we  pretrain the model on ImageNet-1K, and then fine-tune the pretrained backbone on various kinds of other datasets with same protocols and optimization settings  in~\cite{DINO,iBOT}.  Table~\ref{table:transfer} summarizes the classification accuracy, in which ``Sup." denotes the setting where we pretrain the backbone on ImageNet-1K in a supervised manner and then fine tune backbone on the corresponding dataset.   Table~\ref{table:transfer} shows our Mugs surpasses SoTAs on the first five datasets and achieves comparable accuracy on the Car dataset.

\vspace{-1.3em}
\subsubsection{Object detection \& Instance  segmentation.}  
Now we evaluate Mugs on object detection and  instance segmentation on  COCO~\cite{coco}, in which both tasks emphasize the ability to locate and discriminate objects. 
For  fairness, we use the same protocol in~\cite{iBOT} which builds on  Cascade Mask R-CNN \cite{cascadercnn,maskrcnn} to produce bounding boxes and instance masks simultaneously. See optimization settings in Appendix~\ref{moreexp}. Besides SSL approaches, \eg~MoBY~\cite{moby}, we also compare supervised baselines, Swin-T/7~\cite{Swin}  with similar model size as ViT-S/16.   Table~\ref{tab:objectdetection} shows  that on  detection,  Mugs achieves 49.8 AP$^\mathrm{b}$ and makes 0.4 AP$^\mathrm{b}$ improvement over the runner-up, \ie~iBOT.  Fig.~\ref{illustration_attention} (b) shows that Mugs can accurately  locate and classify objects  in COCO.  For instance segmentation,  Mugs also surpasses all compared baselines and improves 0.4 AP$^\mathrm{m}$ over the best baseline. 

\begin{figure*}[t]
	\setcounter{figure}{3}
	\begin{center}
		\setlength{\tabcolsep}{0.8pt}  
		\begin{tabular}{ccc  cc cc }
			\includegraphics[width=0.1\linewidth]{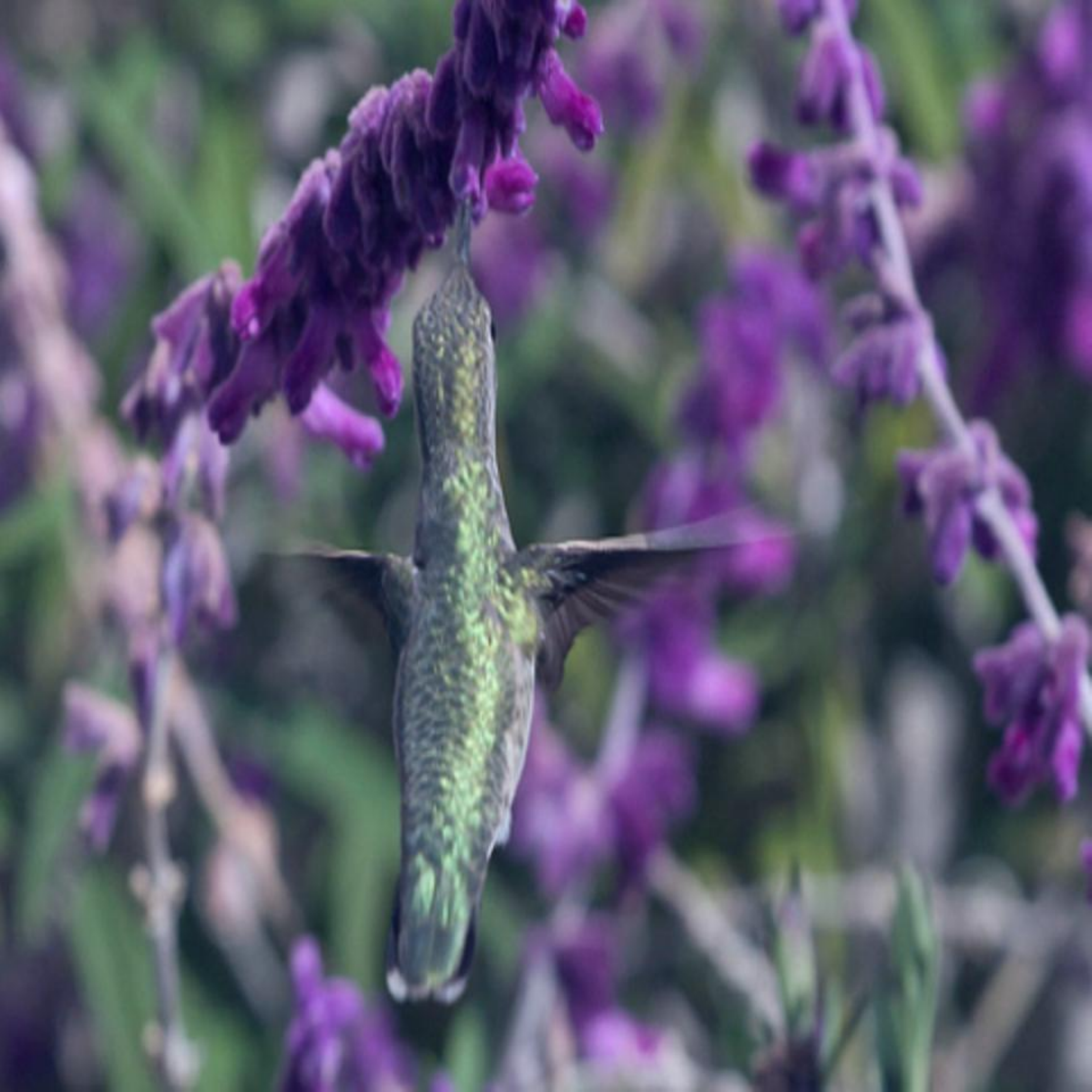}& \includegraphics[width=0.1\linewidth]{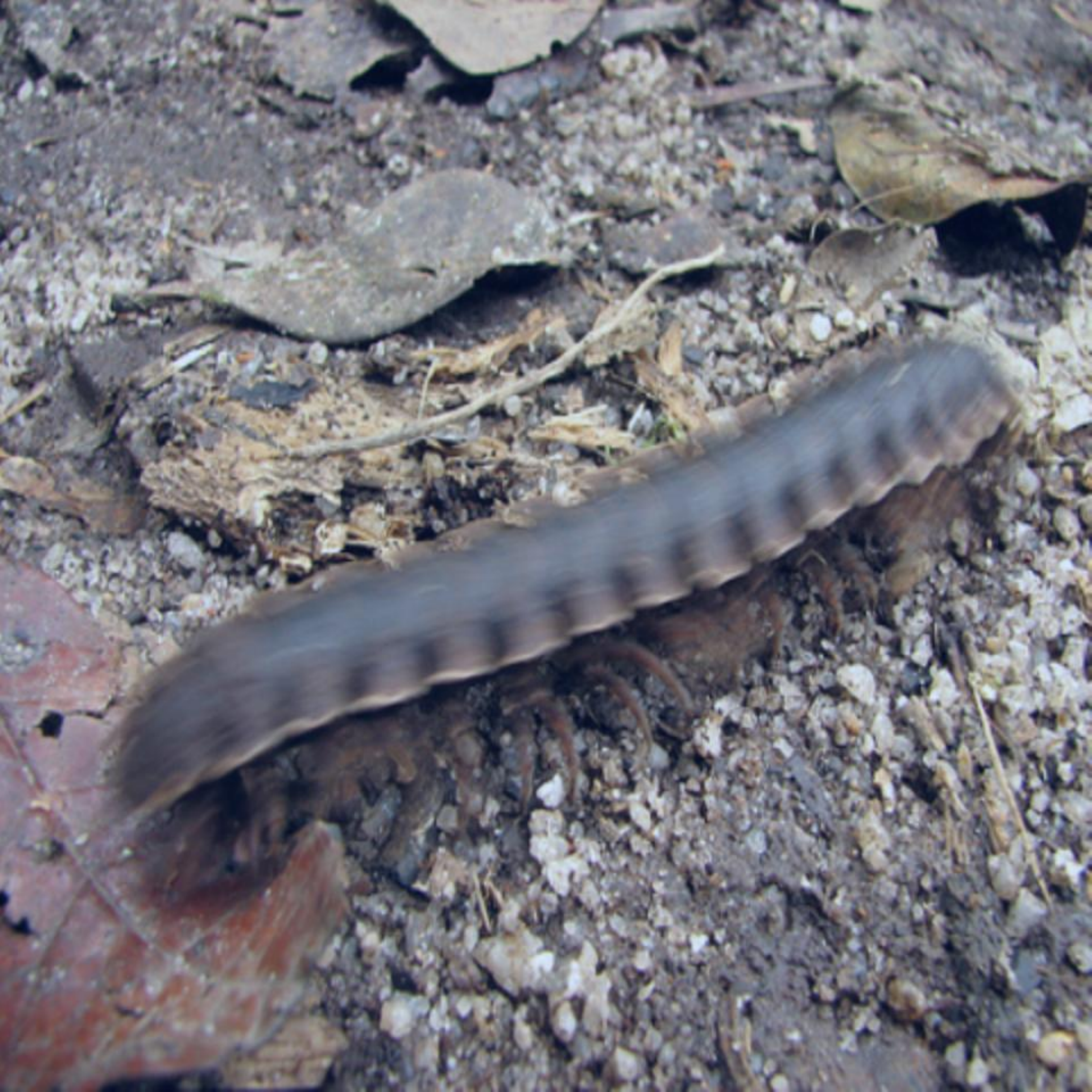} & 
			\includegraphics[width=0.1\linewidth]{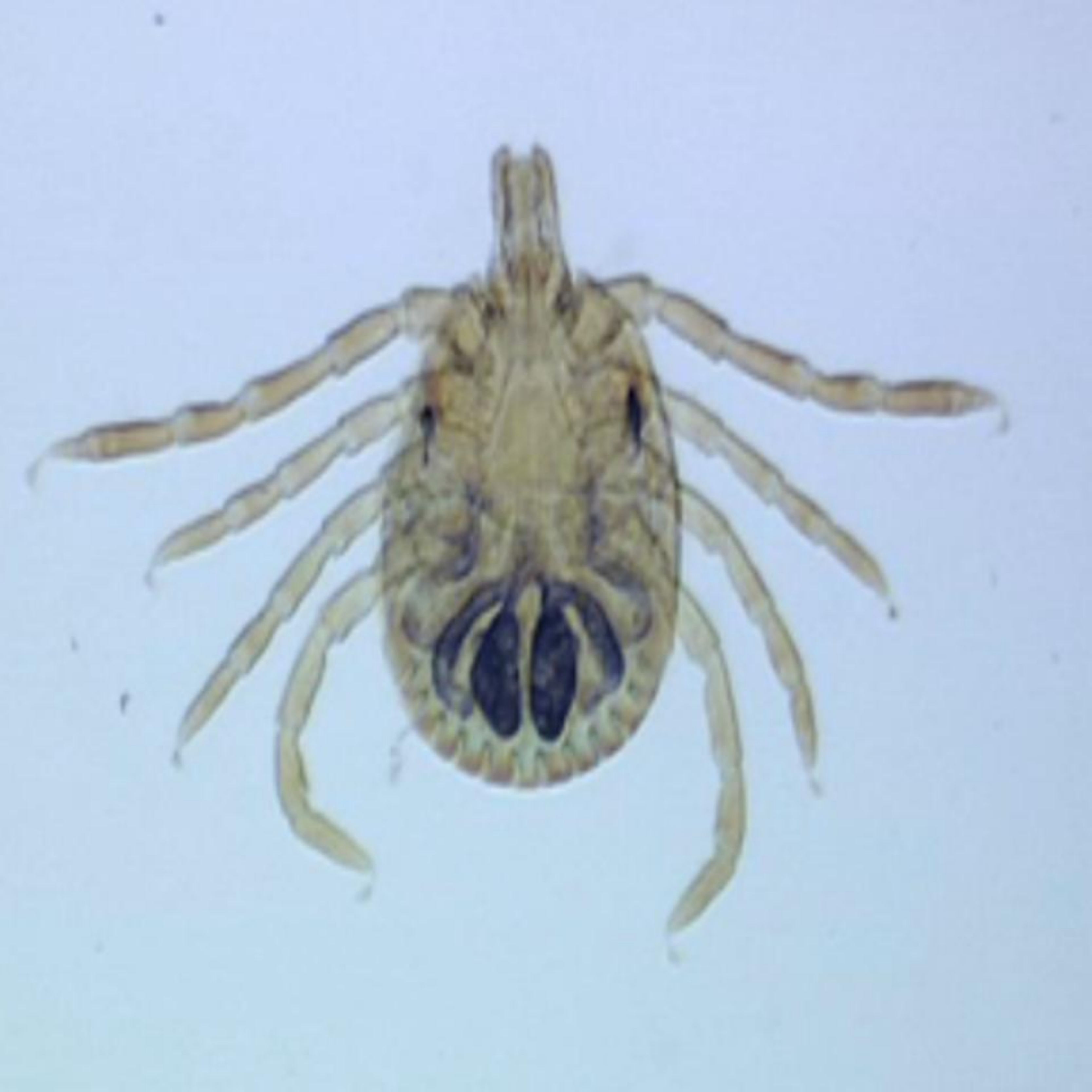}
			\hspace{0.9em}\quad& 
			\includegraphics[width=0.15\linewidth]{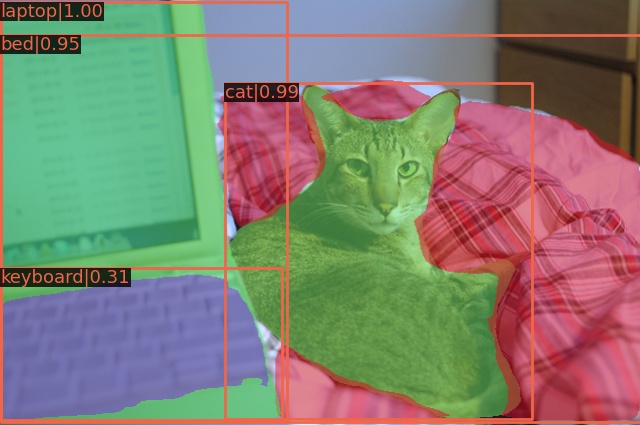} &
			\includegraphics[width=0.15\linewidth]{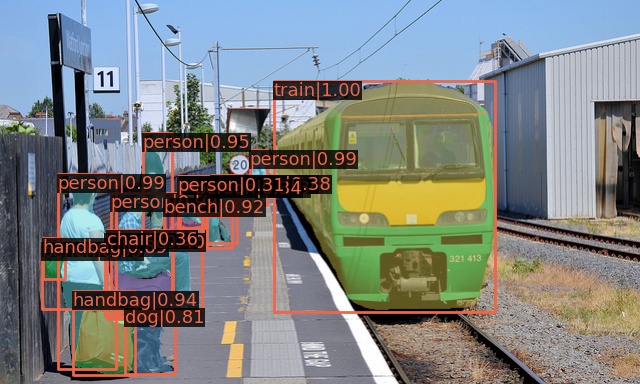} \hspace{0.9em}\quad & 
			\includegraphics[width=0.15\linewidth]{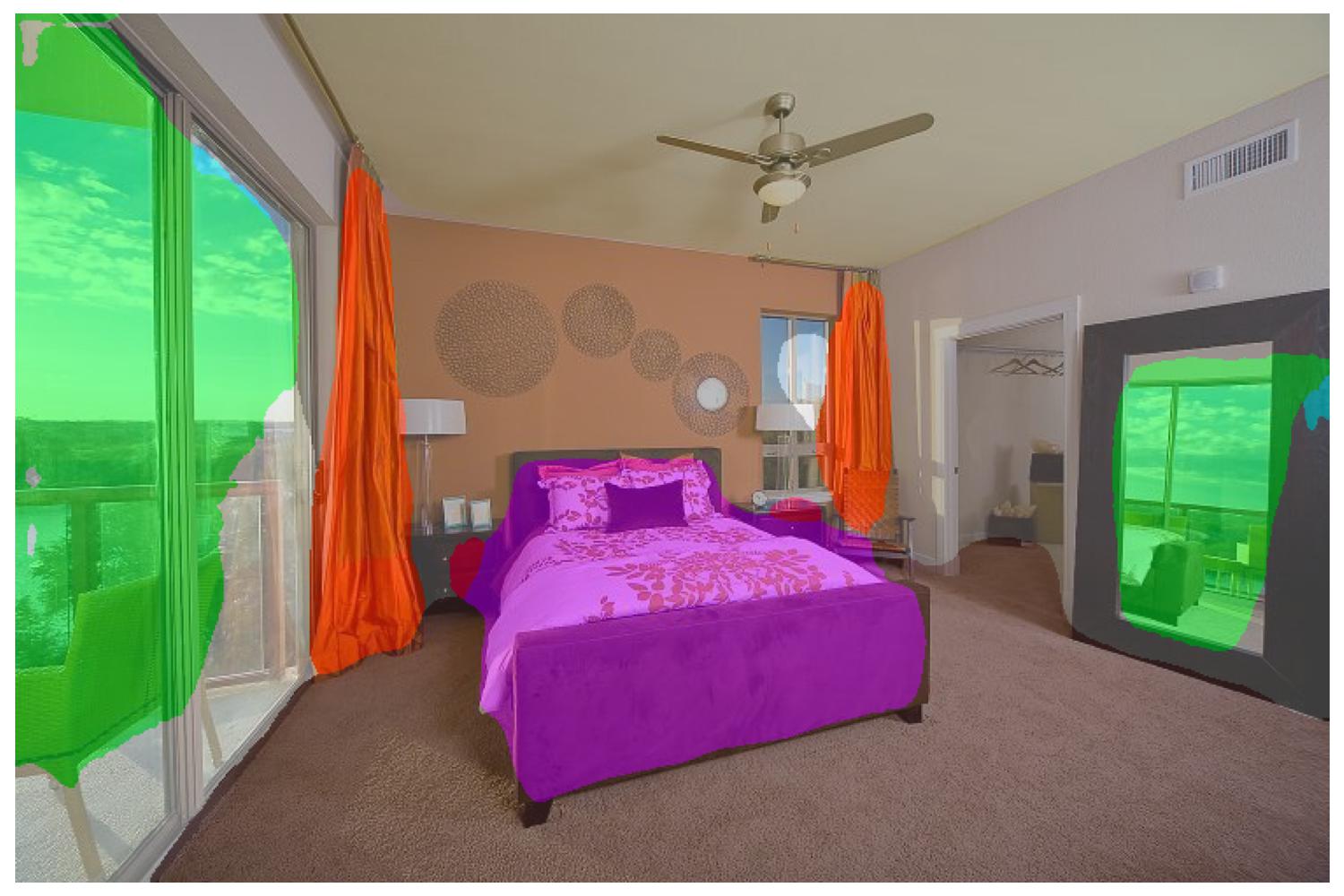} & 
			\includegraphics[width=0.15\linewidth]{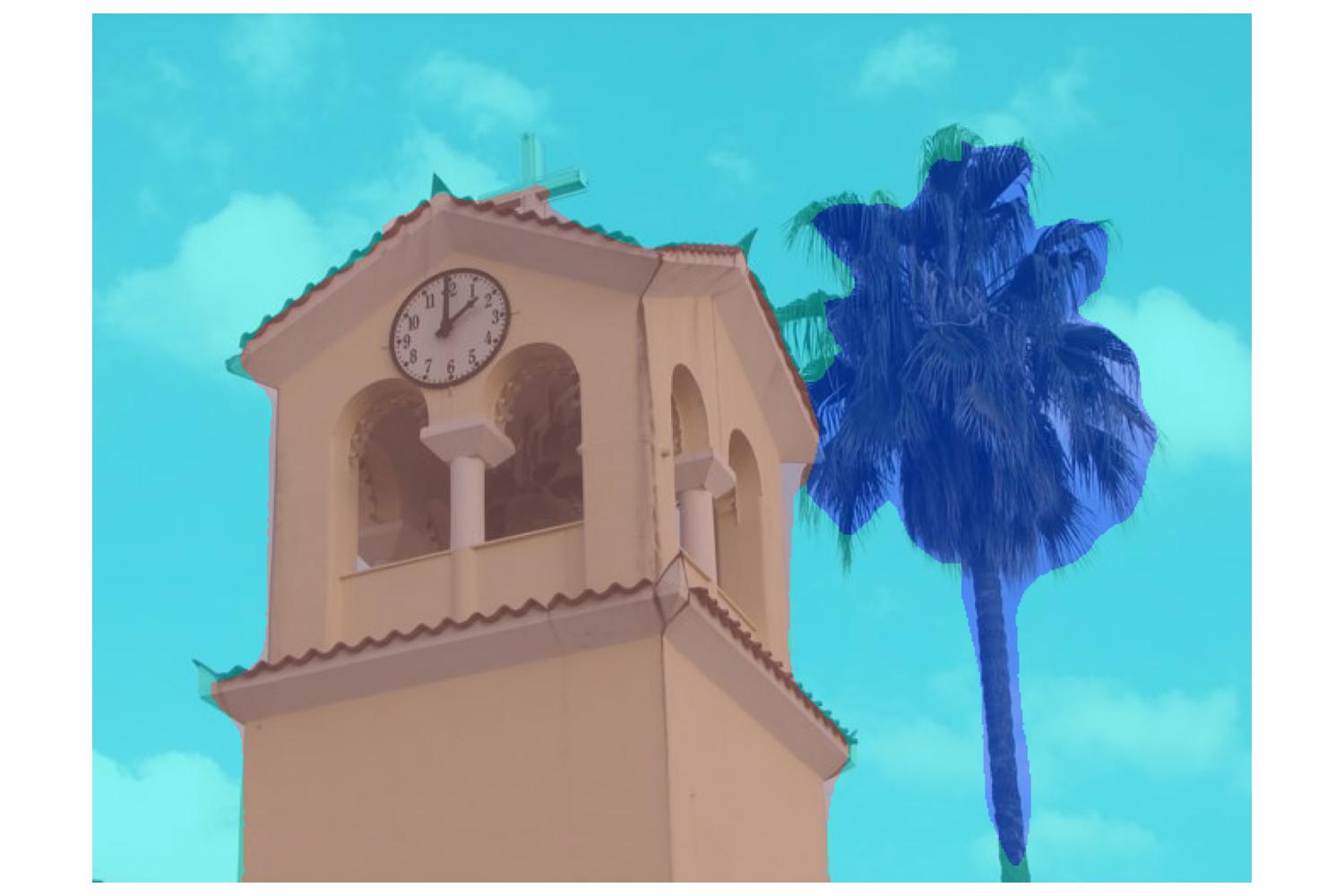} \\
			
			\includegraphics[width=0.1\linewidth]{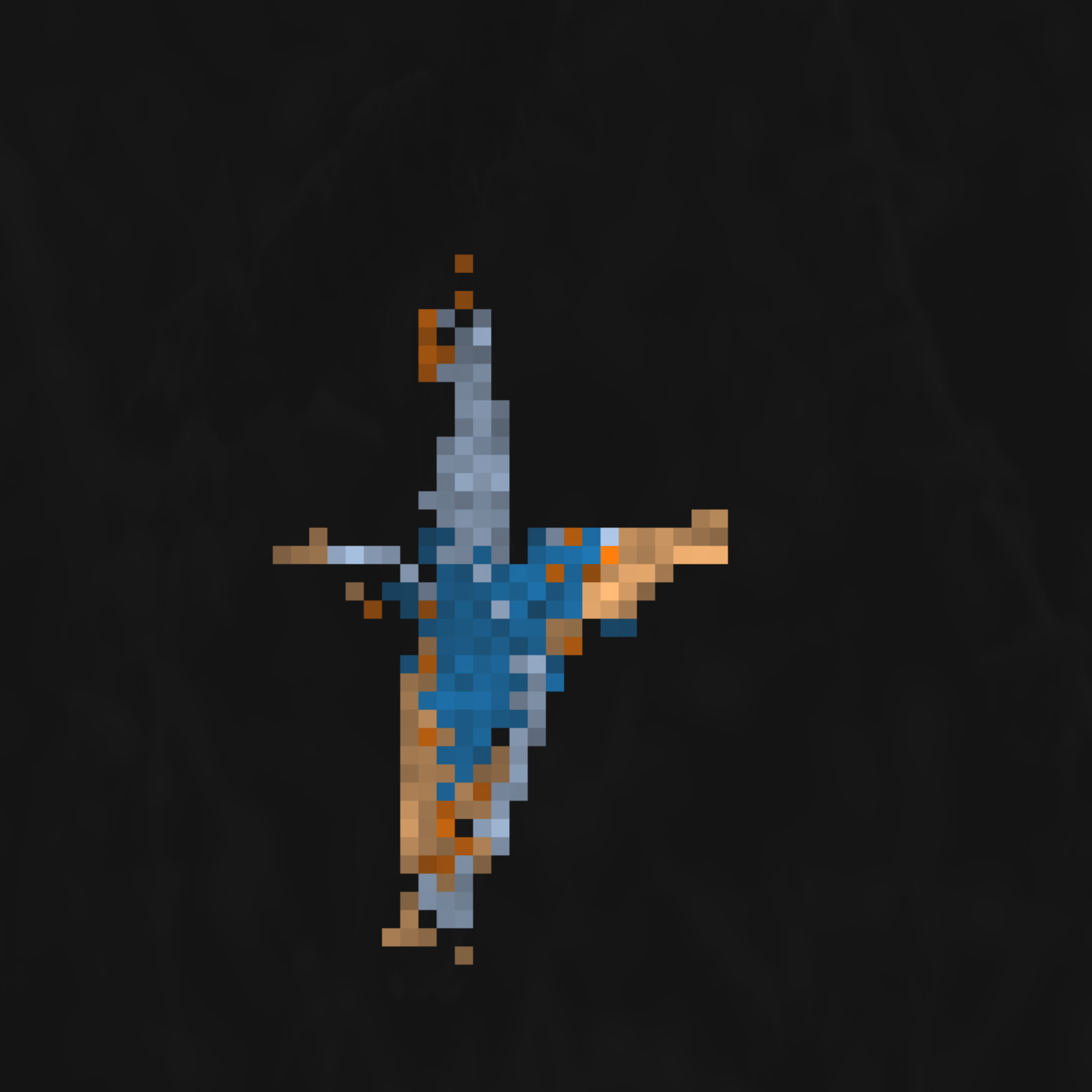} & 
			\includegraphics[width=0.1\linewidth]{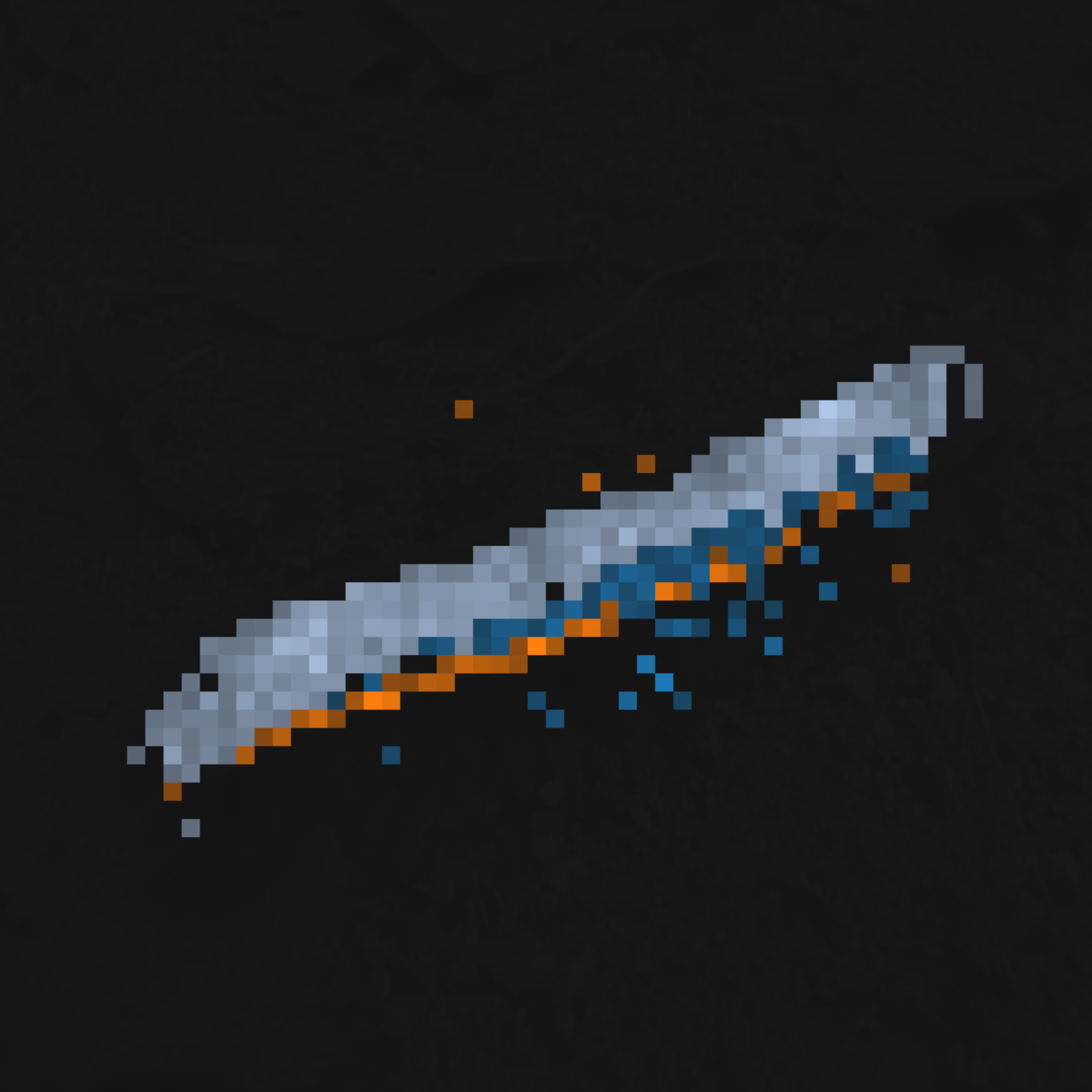}& 
			\includegraphics[width=0.1\linewidth]{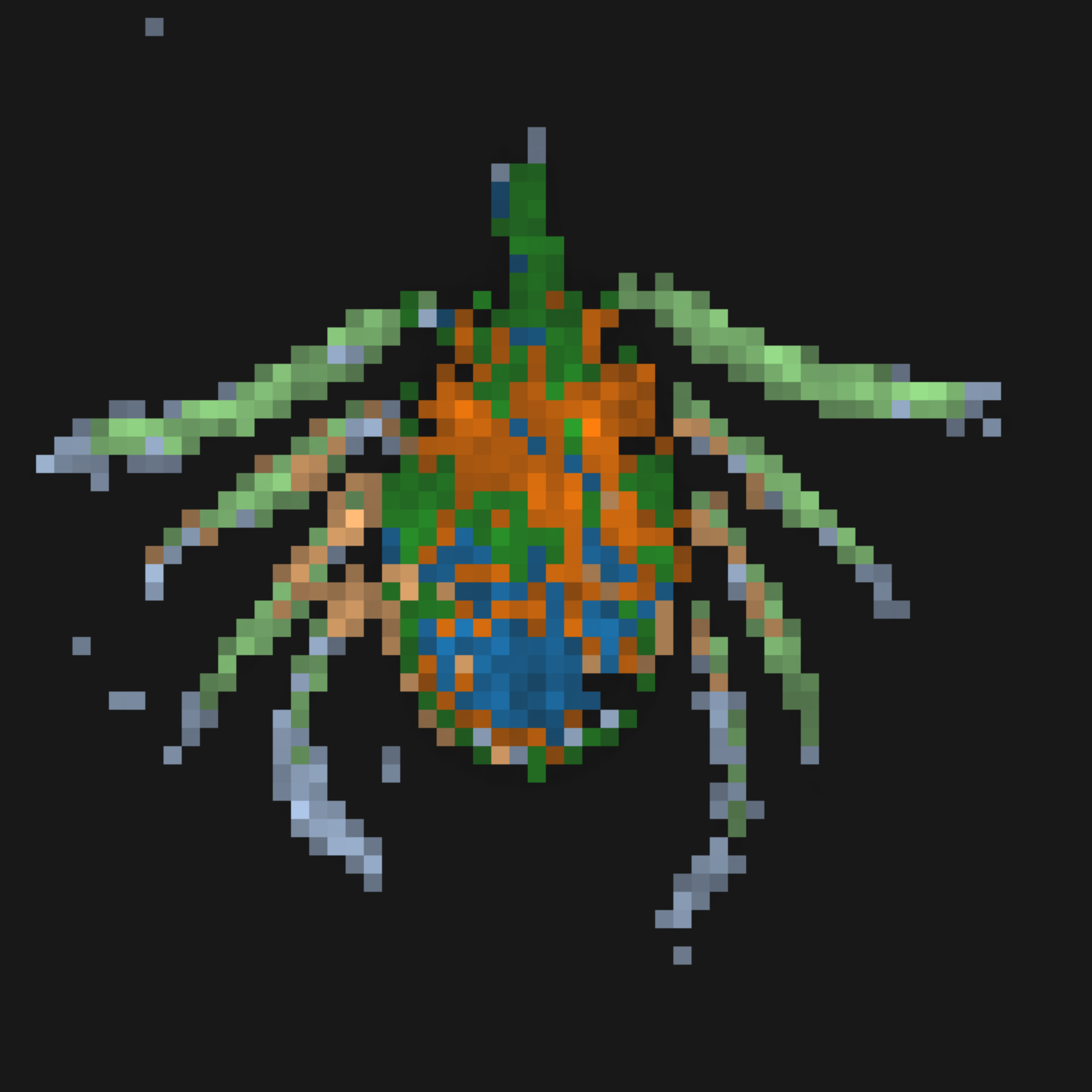} 
			\hspace{0.9em}\quad& 
			\includegraphics[width=0.135\linewidth]{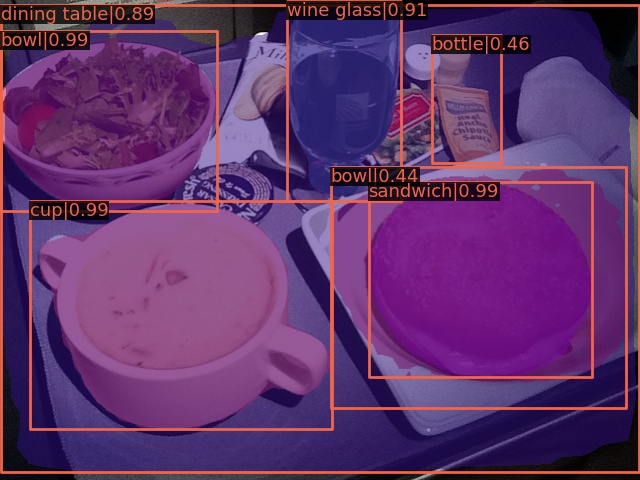}  &
			\includegraphics[width=0.135\linewidth]{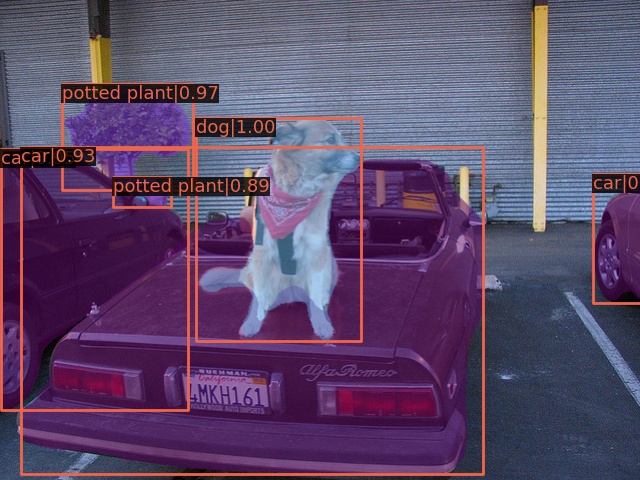} \hspace{0.9em} \quad& 
			\includegraphics[width=0.15\linewidth]{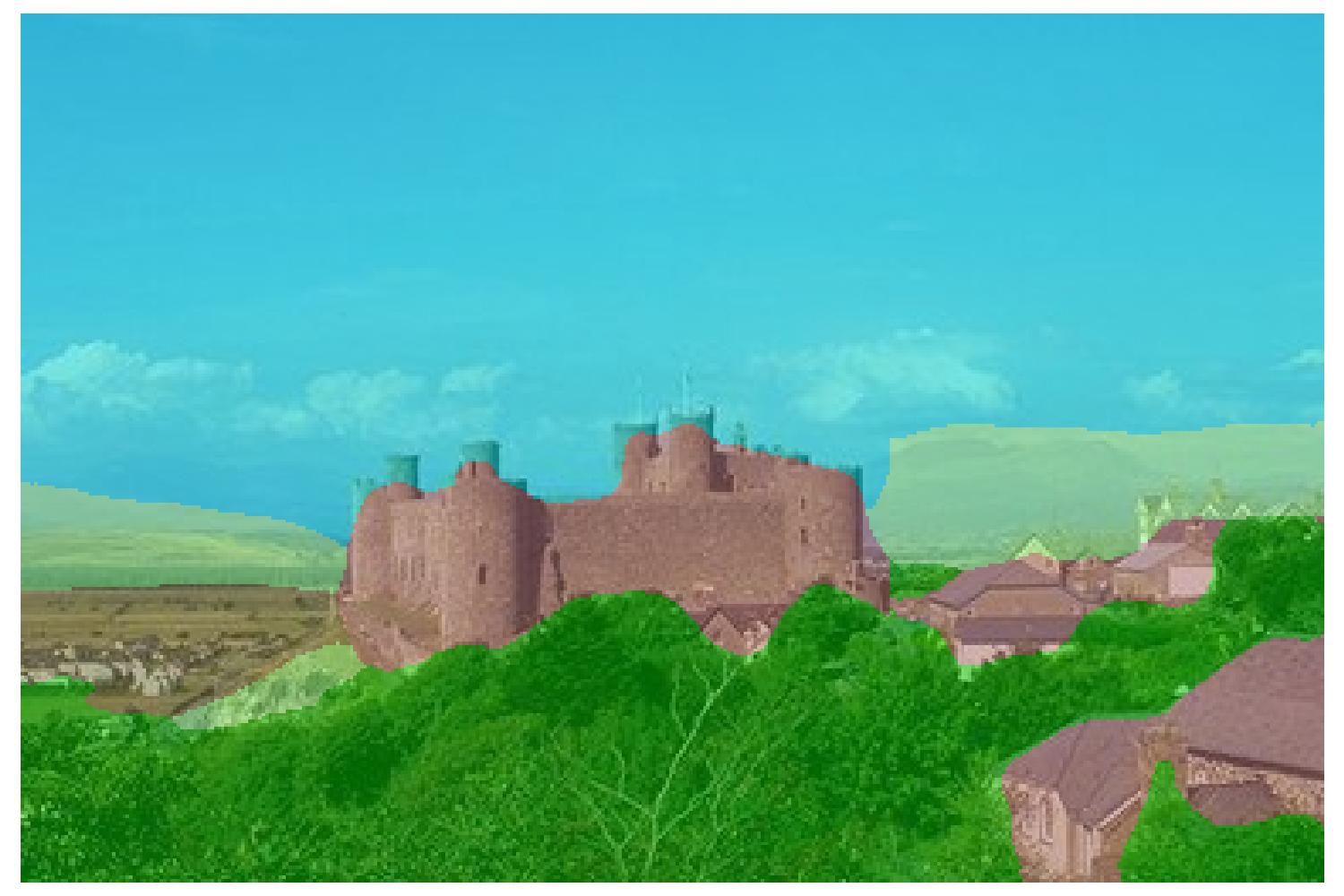} &
			\includegraphics[width=0.15\linewidth]{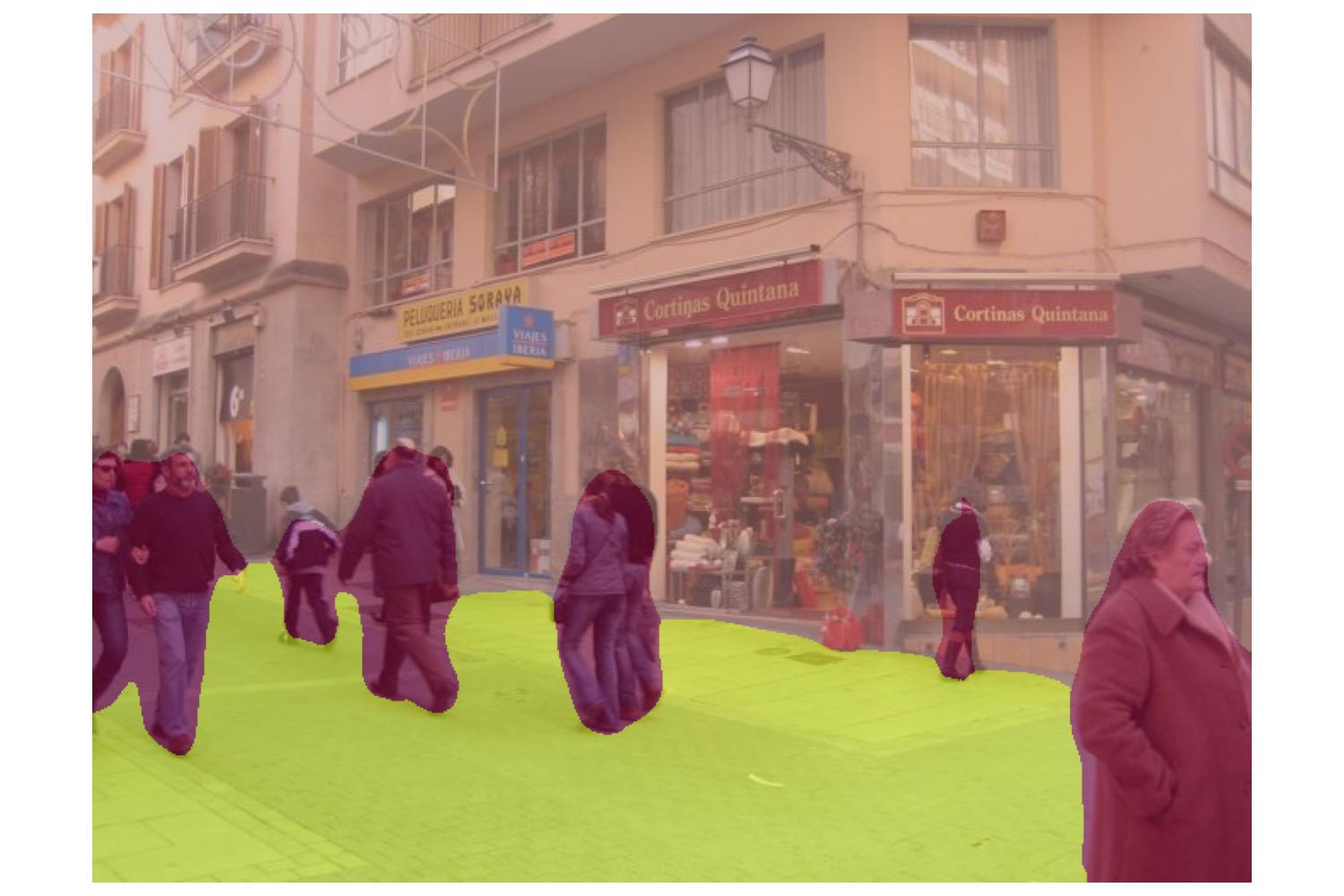} \vspace{-0.3em}\\ 
			\multicolumn{3}{c}{\scriptsize{(a) attention visualization $\quad \ \ $}} &  \multicolumn{2}{c}{\hspace{-1.8em} \scriptsize{(b) object detection}} & \multicolumn{2}{c}{\scriptsize{(c) semantic segmentation}}
		\end{tabular}
	\end{center}
	\vspace{-1.9em}
	\caption{Visualization of pretrained ViT-B/16 (a) and ViT-S/16 (b) \& (c)  by Mugs. 
		See more examples  in Appendix. Best viewed in $3\times$ sized color pdf. }
	\label{illustration_attention}
	\vspace{-0.9em}
\end{figure*}

\begin{table}[tbp]
	\begin{minipage}[c]{.6\linewidth}
		\caption{\!\!\textbf{Object\! detection}\! (Det.)\! \&\! \textbf{instance\! segmentation}\!  (ISeg.)\! on\! COCO\! \&\!  \textbf{semantic\! seg.}\! (SSeg.)\! on\! ADE20K.
		}
		\label{tab:objectdetection}
		\centering
		\setlength{\tabcolsep}{1.2pt} 
		\renewcommand{\arraystretch}{0.85}
		{ \fontsize{7.8}{2}\selectfont{
				\begin{tabular}{l  lcccc }
					\toprule
					& \multirow{2}{*}{\textbf{Arch.}} & \multirow{2}{*}{\textbf{Param.}} & \textbf{Det.} & \textbf{ISeg.} & \multicolumn{1}{c}{\textbf{SSeg.}} \\
					\cmidrule(lr){4-4}\cmidrule(lr){5-5}\cmidrule(lr){6-6}
					& & & AP$^\mathrm{b}$ & AP$^\mathrm{m}$ & mIoU \\
					\toprule
					{Sup.}~\cite{iBOT} &  {Swin-T} &  {29} &  {48.1} & {41.7} & {44.5} \\
					MoBY~\cite{moby} & Swin-T & 29 & 48.1 & 41.5 & 44.1 \\
					{Sup.} ~\cite{iBOT}&  {ViT-S/16} &  {21} &  {46.2} &  {40.1} &  {44.5} \\
					iBOT~\cite{iBOT} & ViT-S/16 & 21 &   49.4 &   42.6 &   45.4 \\
					\textbf{Mugs \scriptsize{(ours)}}  &  ViT-S/16 & 21  &  \bf 49.8 &    \bf 43.0 &    \bf 47.4 \\ 
					\bottomrule
		\end{tabular}}}
	\end{minipage}
	\hspace{0.25cm}
	\begin{minipage}[c]{.35\linewidth}
		\caption{ \textbf{Video object segmentation} with ViT-B/16 on the DAVIS-2017 video  dataset.}
		\label{tab:retrival}
		\centering
		\setlength{\tabcolsep}{1.2pt} 
		\renewcommand{\arraystretch}{3.7}
		{ \fontsize{7.8}{3}\selectfont{
				\begin{tabular}{lccccccc}
					\toprule
					& $(\J \& \F)_m$ & $\J_m$ & $\F_m$  \\
					\toprule
					DINO~\cite{DINO}   & 62.3 & 60.7 & 63.9 \\
					iBOT~\cite{iBOT}  & 62.4 & 60.8 & 64.0		 \\
					\textbf{Mugs}  & \bf 63.1 & \bf 61.4 & \bf 64.9	 \\ 
					\bottomrule
		\end{tabular}}}
	\end{minipage}%
	\vspace{0.1em}
\end{table}

\vspace{-1.2em}
\subsubsection{Semantic segmentation.} We transfer the pretrained model to semantic segmentation task on the ADE20K dataset \cite{ade20k}.  
Following~\cite{iBOT},   we
stack the task layer in UPerNet \cite{upernet} and  fine-tune the whole  backbone.   Table~\ref{tab:objectdetection} reports the mean intersection over union (mIoU)  on  all semantic categories.  Mugs consistently outperforms 
the compared SoTAs by making significant 2.0 mIoU improvement.   Fig.~\ref{illustration_attention} (c) shows that Mugs can  capture the  object shape  accurately.

\vspace{-1.2em}
\subsubsection{Video object segmentation} via nearest neighbor retrieval. Besides images, we further evaluate the transferability of the frozen pretrained  features on videos. 
Following DINO, we find nearest neighbors to segment objects in the video, since one can propagate segmentation masks via retrieving nearest neighbor between consecutive video frames~\cite{philbin2008lost}.  Table~\ref{tab:retrival} reports the mean region similarity $\J_m$  and mean contour-based accuracy  $\F_m$ on the DAVIS-2017 video segmentation dataset~\cite{pont20172017} by using ViT-B/16. 
One can see that Mugs enjoys better feature transferability than  DINO and iBOT even for video segmentation.

\vspace{-1.0em}
\subsection{Ablation Study} \label{AblationStudy}
\vspace{-0.4em}
Here we investigate the effects of each granular supervision in Mugs. Specifically, we train Mugs for 1,00 epochs on ImageNet-1K and report the linear probing accuracy in Table~\ref{table:ablation}. One can observe that by independently removing each granular supervision, namely,  the instance, local-group  and  group supervision, the performance of Mugs degenerates, which shows the benefit of each granular supervision, especially for the local-group supervision.

Next, we compare Mugs with DINO and iBOT under different augmentations and also show the effects of augmentations. Specifically, for the augmentation \scalebox{0.85}{$\Ts$}  in student network of  Mugs/DINO/iBOT, we implement it by  strong  or weak augmentation mentioned at the beginning of Sec.~\ref{exp}; for augmentation \scalebox{0.85}{$\Tt$} in teacher, we always use weak augmentation. See more implementation details in Appendix~\ref{moreexp}, especially for iBOT. We pretrain all methods for 1,00 epochs on ImageNet-1K. Table~\ref{table:strongaugmentation} shows  four observations.  
1) Under weak or strong augmentation, Mugs consistently outperforms  DINO and iBOT. 
2) For all three methods, strong augmentation slightly improves their performance under weak augmentation, showing the effectiveness of our strong augmentation technique on ViTs.   
3) Mugs using weak augmentation  surpasses iBOT that  adopts both  weak augmentation and random mask augmentation.  
4) Under  weak  augmentation, Mugs  improves 1.5\% over DINO which means it is the multi-granular supervisions of Mugs that contributes this 1.5\% improvement. Then by using strong augmentation,  Mugs  surpasses DINO using weak augmentation by 2.2\%, showing strong augmentation only contributes 0.7\%  improvement over DINO.
So compared with the strong augmentation technique, the multi-granular supervision framework of Mugs   largely contributes to Mugs and is the key factor  to the significant improvement of Mugs over DINO and iBOT.

Finally, we evaluate Mugs without multi-crop augmentation, \ie~only using two crops of size 224$\times$224 for pretraining. Table~\ref{tab:nomulticrop} in Appendix~\ref{moreResults} shows that  Mugs  also surpasses  the SoTAs, including  DINO and iBOT,  on ViTs under the same setting, which also demonstrates the  superiority of Mugs.

\begin{table}[t]
	\begin{center}
		\caption{Effects of the three granular supervisions  in Mugs to the linear probing accuracy (\%) on ImageNet-1K. $\LL_{\text{instance}}$,   $\LL_{\text{local-group}}$ and  $\LL_{\text{group}}$ respectively denote the instance, local-group and group discrimination supervision. }
		\label{table:ablation}
		\setlength{\tabcolsep}{4.5pt} 
		\renewcommand{\arraystretch}{1.0}
		{ \fontsize{8.3}{3}\selectfont{
				\begin{tabular}{c||c|c|c|c}
					\toprule
					{\bf{Method}} & Mugs & Mugs w/o $\LL_{\text{instance}}$ &  Mugs w/o  $\LL_{\text{local-group}}$ &  Mugs w/o  $\LL_{\text{group}}$ \\
					{\bf{Acc (\%)}}  & 76.4 &  75.8 & 75.3  & 75.7\\ 
					\bottomrule
		\end{tabular}}}
	\end{center}
	\vspace{-1.9em}
\end{table}

\begin{table}[t]
	\begin{center}
		\caption{Augmentation effects to linear probing accuracy (\%) on ImageNet-1K. $\dagger$ denotes  that we replace vanilla augmentation in the method and run this variant. }
		\label{table:strongaugmentation}
		\setlength{\tabcolsep}{6.5pt} 
		\renewcommand{\arraystretch}{1.0}
		{ \fontsize{8.3}{3}\selectfont{
				\begin{tabular}{ ccc | ccc | c}
					\toprule
					\multicolumn{3}{c|}{ {weak aug.}} &	\multicolumn{3}{c|}{ {strong aug.}}  & 	weak aug.+random mask  \\ 
					DINO & iBOT{$^\dagger$} & Mugs & DINO{$^\dagger$}  & iBOT{$^\dagger$}  & Mugs & iBOT \\
					\hline 
					74.2 & 74.9 & 75.7 & 74.7 & 75.4 & 76.4 & 75.3 \\
					\bottomrule
		\end{tabular}}}
	\end{center}
	\vspace{-0.8em}
\end{table}

\vspace{-1.1em}
\section{Conclusion} 	
\vspace{-0.3em}
In this work, we propose Mugs to learn multi-granular features via three complementary granular supervisions: instance discrimination supervision (IDS),  local-group discrimination supervision (LGDS),  
and group discrimination supervision (GDS). IDS  distinguishes different instances to learn  fine-grained features. LGDS considers the local-group around an instance and then discriminates  different local-groups to extract higher-level fine-grained features. GDS clusters similar  samples and local-groups  into one cluster to  capture coarse-grained global group semantics.   Experimental results have testified the advantages of Mugs.

\clearpage
\section*{Acknowledgement} 
\vspace{-0.3em}
We would like to thank NVIDIA AI Tech Center (NVAITC)  for the support of partial computational resources, and also thank Terry Jianxiong Yin (NVAITC) and Qingyi Tao (NVAITC) for their  some GPU technology supports.	
\bibliographystyle{splncs04}
\bibliography{referen}

\newpage

\appendix

\title{Mugs: A Multi-Granular Self-Supervised  Learning Framework (Supplementary File)}

\titlerunning{Mugs: A Multi-Granular Self-Supervised  Learning Framework} 
\authorrunning{Preprint}

\author{\normalsize{Pan Zhou$^{1*}$}   \  \normalsize{Yichen Zhou$^{1*}$}   \   \normalsize{Chenyang Si$^{1*}$}    \    \normalsize{Weihao Yu$^{1}$}    \   \normalsize{Teck Khim Ng$^{2}$}   \    \normalsize{Shuicheng Yan$^{1}$}\\
	\institute{$^{1}$ Sea AI Lab, Singapore \qquad  $^{2}$ National University of Singapore} 
}
\maketitle

\begin{quote}
	This supplementary document provides more additional experimental results  and the pretraining \& fine-tuning details for  the ECCV'22 submission entitled ``Mugs: A Multi-Granular Self-supervised  Learning Framework''. It is structured as follows. Appendix~\ref{moreResults}  provides  more extra  experimental results, including 1) the comparison among SoTAs without the multi-crop augmentation strategy in Appendix~\ref{multicrop},  2) fine-tuning comparison on ViT-L/16 in Appendix~\ref{finetuning}, 3) more T-SNE clustering visualization results  in Appendix~\ref{attention_Visualization}, 4) more attention visualization results  in Appendix~\ref{attention},  5) more visualization results  on object detection and  segmentation in Appendix~\ref{Objectdetectionsec}.  
	
	Appendix~\ref{moreexp} provides more experimental details  for Sec.~\ref{exp} in   manuscript.  Specifically,  Appendix~\ref{morepretraining} gives more pretraining details, including implementations of weak and strong augmentations,  loss construction under multi-crop setting, hyper-parameter settings  for pretraining and the pretraining cost.   Then Appendix~\ref{1Kdetails} introduces more details  for fine-tuning and semi-supervised learning in Sec.~\ref{resultsonImagenet} in manuscript.  
	Next, in Appendix~\ref{downstreamtasks}, we present more details for downsteam tasks, including transfer learning, object detection \& instance  segmentation, and semantic segmentation. Finally,  Appendix~\ref{appendixAblationStudy} tells us more implementation details of DINO and iBOT under weak and strong agumentation settings to complement Sec.~\ref{AblationStudy} in  manuscript. 
\end{quote}

\appendix

\begin{table}[b!p]
	\setcounter{table}{8}
	\caption{\textbf{Linear probing accuracy (\%) and k-NN accuracy (\%)} on ImageNet-1K without multi-crop augmentation (left) and with multi-crop augmentation (right).  ``Epo" is the effective pretraining epochs adjusted by number of views processed by the models following~\cite{iBOT}.}
	\label{tab:nomulticrop}
	\begin{minipage}[c]{.5\linewidth}
		\centering
		\setlength{\tabcolsep}{3.85pt} 
		\renewcommand{\arraystretch}{1.1}
		{ \fontsize{8.3}{3}\selectfont{
				\begin{tabular}{llcccc}
					\toprule
					\textbf{Method}  & \textbf{Para.} & \textbf{Epo.} & \textbf{Lin.} & \textbf{k-NN}  \\
					\midrule
					{DINO}  & 21 & 3200 & 73.7 & 70.0  \\  
					{iBOT}  & 21 & 3200 &     76.2 &    72.4 \\  
					{\textbf{Mugs}} & 21 & 3200 & \bf \pz{76.9} & \bf  \pz{73.1}   \\
					\bottomrule
		\end{tabular}}}
	\end{minipage}%
	\hspace{0.2cm}
	\begin{minipage}[c]{.5\linewidth}
		\setlength{\tabcolsep}{3.85pt} 
		\renewcommand{\arraystretch}{1.1}
		{ \fontsize{8.3}{3}\selectfont{
				\begin{tabular}{lccccc}
					\toprule
					\textbf{Method}    & \textbf{Para.} & \textbf{Epo.} & \textbf{Lin.} & \textbf{k-NN}  \\
					\midrule
					{DINO} & 21 & 3200 & 77.0 & 74.5  \\ 
					{iBOT}  & 21 & 3200  &   77.9 &   75.2 \\  
					\textbf{Mugs}  & 21 & 3200 &  \textbf{78.9} &  \textbf{75.6}  \\ 
					\bottomrule
		\end{tabular}}}
	\end{minipage}
\end{table}
\section{More Experimental Results}\label{moreResults}

Due to space limitation, we defer more experimental results to this appendix. Here we
first investigate the performance of Mugs without  multi-crop augmentations which is widely used in several representative works, and further compare it with other methods, include iBOT and DINO under the same setting.  Then we present more visualization results, including T-SNE clustering visualization, attention visualization of multi-heads in ViT, and object detection and segmentation visualization.  We hope these visualization results can help readers intuitively understand the learnt features by Mugs.

\subsection{Comparison w/o and w/ Multi-Crop Augmentation}\label{multicrop}

Here we
first investigate the performance of Mugs without the multi-crop augmentation which is widely used in several representative works, and further compare it with other SoTA methods, include iBOT and DINO under the same setting.  Specifically,  for  Mugs without multi-crop augmentation, it~only uses two 224-sized crops for pretraining. The left table in Table~\ref{tab:nomulticrop}  reports the  results of all compared methods without multi-crop augmentation, while the right one summarizes  the  results under  multi-crop augmentation setting. By comparison, one can observe that without multi-crop augmentation, Mugs still consistently achieves the highest accuracy under both linear probing setting and KNN setting. Specifically, Mugs improves  the runner-up, namely iBOT, by respectively   \pz{$0.8\%$}  and \pz{$0.5\%$}  under linear probing and KNN evaluation settings. More importantly, we can observe that Mugs without multi-crop augmentation even achieves very similar results as  DINO with multi-crop augmentation. All these results are consistent with those results in Table~\ref{tablelinear} in the manuscript,  and well demonstrate the superiority of Mugs over previous state-of-the-arts.

\begin{table}[t]
	\caption{\textbf{Fine-tuning}  classification accuracy (\%)  on  ImageNet-1K. All methods are pretrained on ImageNet-1K. ``Epo." is the effective pretraining epochs adjusted by number of views processed by the models following~\cite{iBOT}.\vspace{-1.0em}}
	\label{tablefinetune2}
	\begin{center}
		\setlength{\tabcolsep}{9.6pt} 
		\renewcommand{\arraystretch}{0.85}
		{ \fontsize{8.3}{3}\selectfont{
				\begin{tabular}{c | l | c c cc  c c }
					\toprule
					&	\multirow{2}{*}{\textbf{Method}}  &  \multicolumn{2}{c}{\textbf{ViT-L/16}} \\
					&		\textbf{}  & {Epo.} & {Acc. (\%)}     \\
					\midrule 
					&	Supervised~\cite{DeiT}    & --- & 83.1 \\
					\hline
					
					&  		BEiT~\cite{BEiT}      & 800 & 85.2  \\
					reconstruction 	&MAE~\cite{MAE}      &  1600 &  85.9 \\
					
					&	data2vec~\cite{Alexei2022datavec}     & 1600 & \textbf{86.6} \\
					\hline
					&	DINO~\cite{DINO}    & --- & --- \\
					contrastive or		&	iBOT~\cite{iBOT}     &  1000  & 84.8 \\
					clustering	&	MoCo-v3~\cite{MoCo-v3}    & 600 & 84.1 \\
					&	\textbf{Mugs \scriptsize{(ours)}}      & 1000 & 85.2  \\   
					\bottomrule
		\end{tabular}}}
	\end{center}
\end{table}

\subsection{Comparison under Fine-tuning Setting}\label{finetuning}
In the manuscript, we already compare Mugs with state-of-the-art approaches on the ViT-S/16 and ViT-B/16 under the fine-tuning setting. Due to limited space,  we defer the comparison among Mugs and others on ViT-L/16  into Table~\ref{tablefinetune2}.  This setting allows us to  optimize the pretrained backbone with a linear classifier.  Following BEiT~\cite{BEiT}, DINO and iBOT, we use AdamW optimizer with layer-wise learning rate decay to train ViT-L for 50 epochs on ImageNet-1K.  
On  ViT-L, Mugs  achieves  85.2\% top-1 accuracy, and surpasses all contrastive learning and clustering learning methods.  One  can also observe that on ViT-L,  most of the reconstruction methods  achieves higher accuracy than constrictive or clustering learning approaches, including iBOT and our Mugs.  There are two possible reasons. Firstly,  the reconstruction methods use much more computations  for pretraining than constrictive or clustering learning approaches. Specifically, the  reconstruction family always use 224$\times$224-sized images to pretrain the  model,  while  constrictive or clustering learning approaches  uses multi-crop augmentations which contains two  224-sized images and ten 96-sized images.  Since ``Epo." in Table~\ref{tablefinetune2} is the effective pretraining epochs adjusted by number of views processed by the models~\cite{iBOT} which means each $96$-sized image equals to one $224$-sized image in terms of  the defined ``epochs",    with the same pretraining  epochs, the computation cost of the reconstruction approaches is much more. Actually, from Table~\ref{tablefinetune2}, the reconstruction methods have much more effective pretraining epochs than constrictive or clustering learning approaches, e.g. 1600 epochs in data2vec v.s. 1000 epochs in iBOT \& Mugs, which further increases the training cost.   Secondly, for large models, using small-sized images, e.g. ten 96-sized images in multi-crop augmentations, may lead to overfitting issue in contrastive or clustering learning approaches.  Specifically, from Table~\ref{tablelinear} in manuscript and Table~\ref{tablefinetune2} here, once can observe that on relatively small models, such as  ViT-S and ViT-B,  SoTA contrastive learning or clustering methods, such as Mugs and iBOT, outperform the  reconstruction methods, even though the formers have much less pretraining cost as mentioned above. But on large models, e.g. ViT-L, the superiority of  SoTA contrastive  or clustering learning methods disappears. One possible reason for these inconsistent observation is that large model needs more pretraining epochs for  learning semantic features, and could suffer from over-fitting problem when using 96-sized crops, since 1) large model is capable to memory all images as demonstrated in many works; and 2) 96-sized crops may contain incomplete semantics of the vanilla image and lead to over-fitting issue, especially under insufficient pretraining epochs.  Note, as explained at the end of Sec.~\ref{relatedwork} in manuscript,  this fine-tuning setting  needs much  higher extra training cost, and also destroys model compatibility for  deployment. Therefore,  in this work,  we do not further push Mugs's limits on the large models which needs huge training cost as the reconstruction methods.

\begin{figure*}[t]
	\setcounter{figure}{4}
	\begin{center}
		\setlength{\tabcolsep}{0.8pt}  
		\begin{tabular}{c}
			\includegraphics[width=1.0\linewidth]{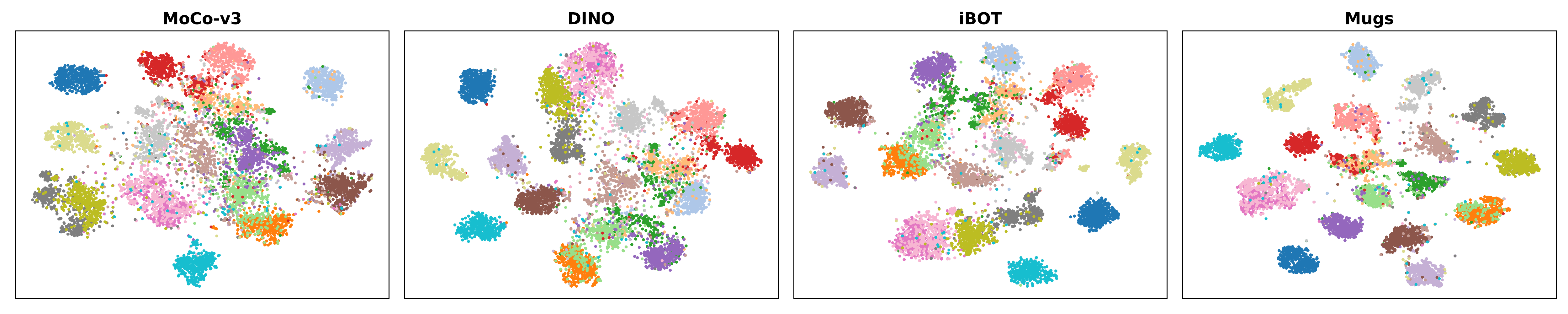}\\
			{ \scriptsize{(a) T-SNE visualization of  19 classes of insects, \eg~beetle, butterfly, stick, and  cricket.}}  \\
			\includegraphics[width=1.0\linewidth]{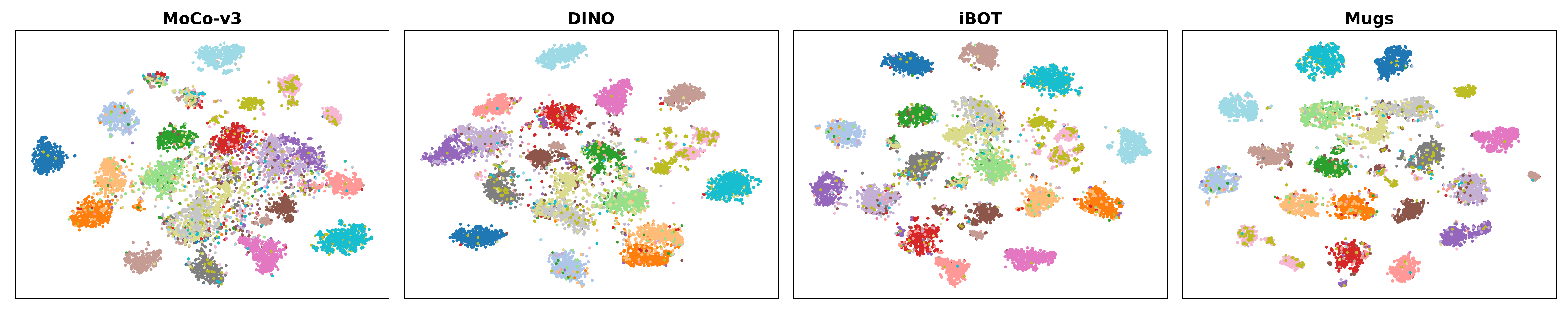}\\
			{ \scriptsize{(b) T-SNE visualization of  20 classes of various monkeys, \eg~gibbon, siamang, patas, and gorilla.}}  \\ 
			\includegraphics[width=1.0\linewidth]{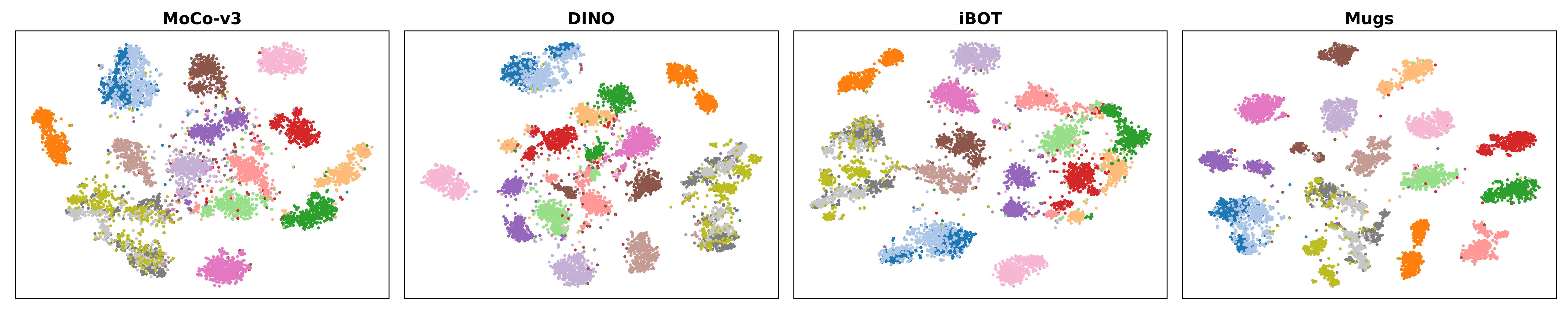}\\
			{ \scriptsize{(c) T-SNE visualization of  17 classes of various birds, \eg~junco,  robin, jay, cock, and  ostrich.}}  \\
		\end{tabular}
	\end{center}
	\vspace{-1.5em}
	\caption{More T-SNE visualization of the  learned features by ViT-B/16 trained by our Mugs.   \textbf{Best viewed in  color pdf file.} }
	\label{illustration_clustering2}
\end{figure*}

\subsection{More T-SNE Visualization Results}\label{attention_Visualization}
Same with  Fig.~\ref{illustration_clustering} in the manuscript, here we use T-SNE~\cite{van2008visualizing} to reveal the feature differences among MoCo-v3, DINO, iBOT, and Mugs in Fig.~\ref{illustration_clustering2}.  By comparison, Mugs often can scatter the samples from different classes more separately, while keeping the samples in the same class close in the feature space.  This could means that our Mugs can better  distinguish different classes than  MoCo-v3, DINO and iBOT, and thus shows higher performance.  The  potential reason behind  this observation is explained in manuscript. That is, instead of regards the class as a whole, Mugs utilizes its multi-granular supervisions to consider the multi-granular (hierarchical)  data semantic structures and divides the whole class into several clusters for easily discriminating in the pretraining phase. Differently, MoCo-v3, DINO and iBOT ignore the multi-granular semantic  structures and only uses one granular supervision which often could not well handle the challenging classes.

\begin{figure*}[h!tb]
	\begin{center}
		\setlength{\tabcolsep}{0.8pt}  
		\begin{tabular}{c}
			\includegraphics[width=1.0\linewidth]{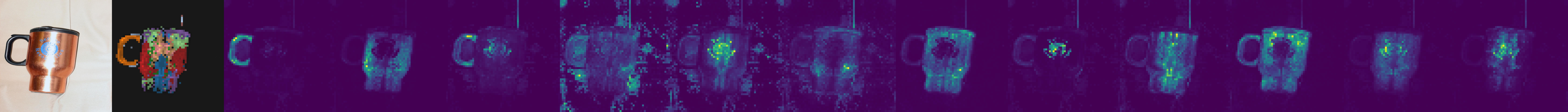}\\
			\includegraphics[width=1.0\linewidth]{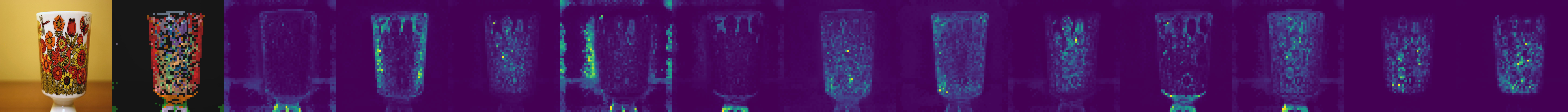}\\  
			\includegraphics[width=1.0\linewidth]{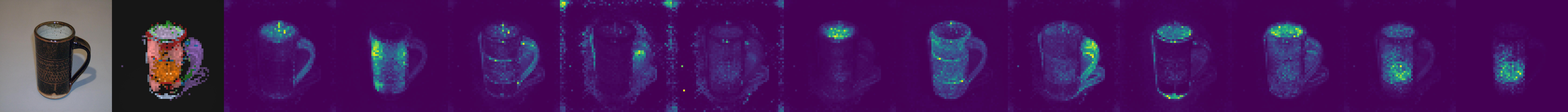}\\ 
			\includegraphics[width=1.0\linewidth]{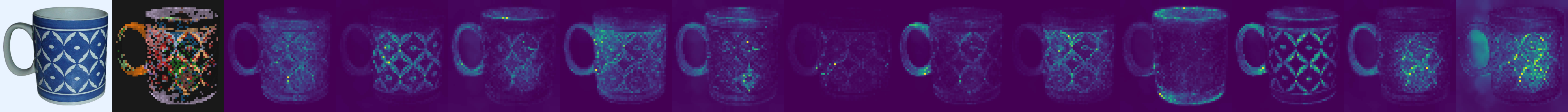}\\ 
			\includegraphics[width=1.0\linewidth]{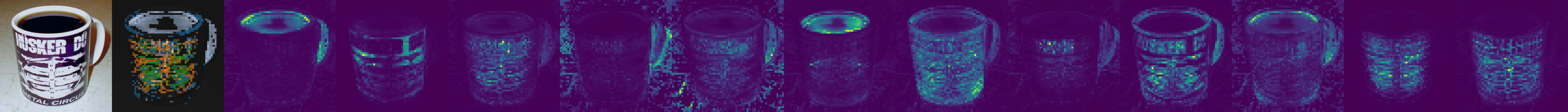}\\  
			\includegraphics[width=1.0\linewidth]{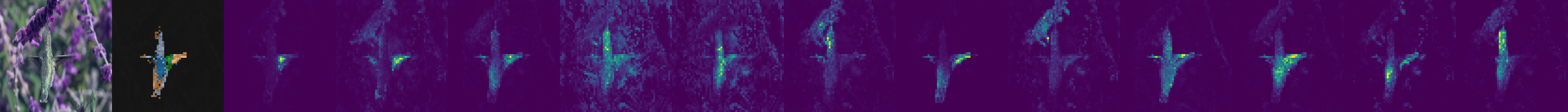}\\
			\includegraphics[width=1.0\linewidth]{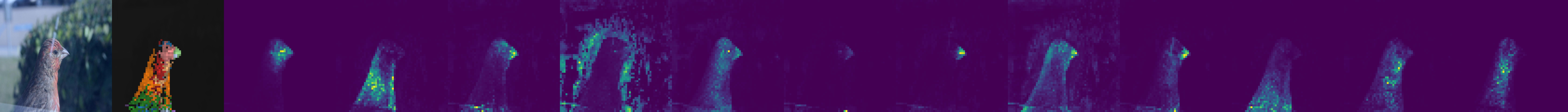}\\
			\includegraphics[width=1.0\linewidth]{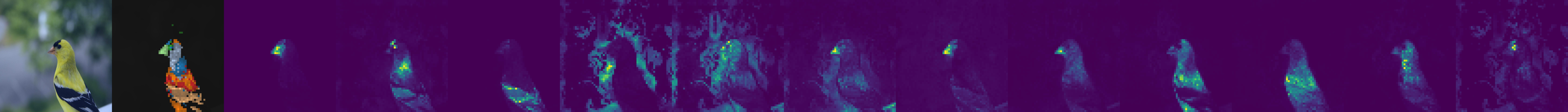}\\
			\includegraphics[width=1.0\linewidth]{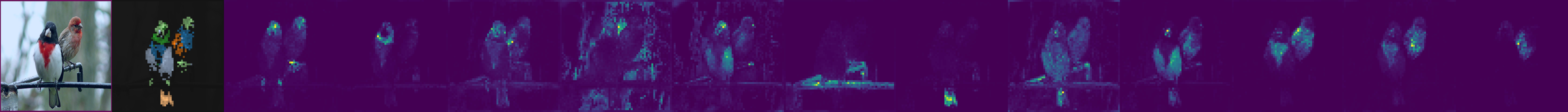}\\
			\includegraphics[width=1.0\linewidth]{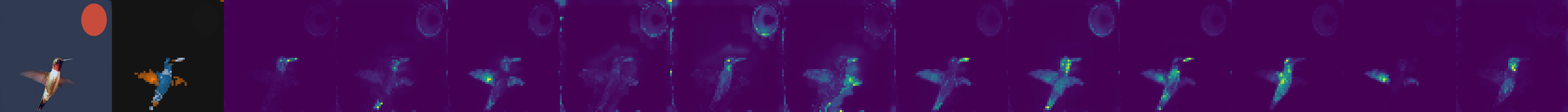}\\
			\includegraphics[width=1.0\linewidth]{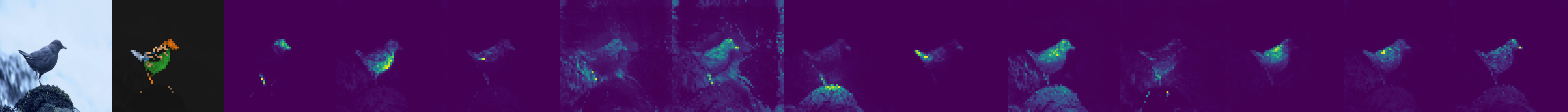}\\
			\includegraphics[width=1.0\linewidth]{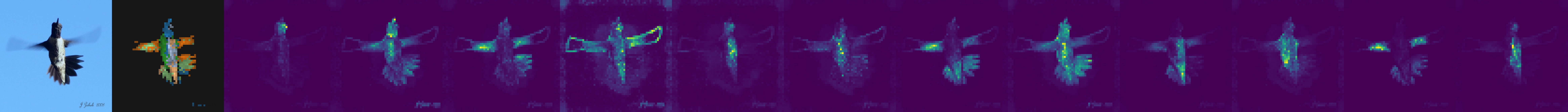}\\ 
			\includegraphics[width=1.0\linewidth]{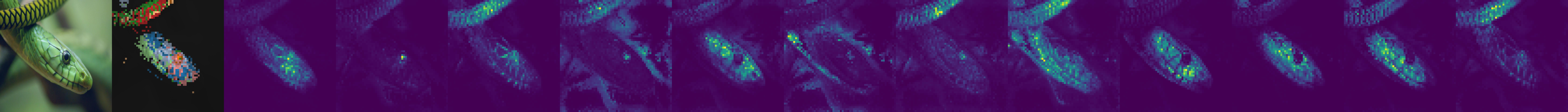}\\
			\includegraphics[width=1.0\linewidth]{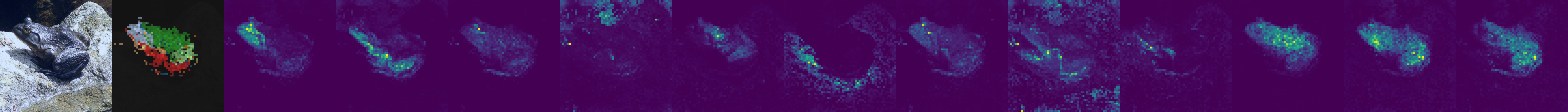}\\ 
			\includegraphics[width=1.0\linewidth]{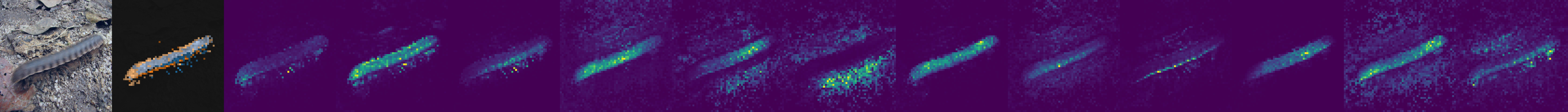}\\ 
			\includegraphics[width=1.0\linewidth]{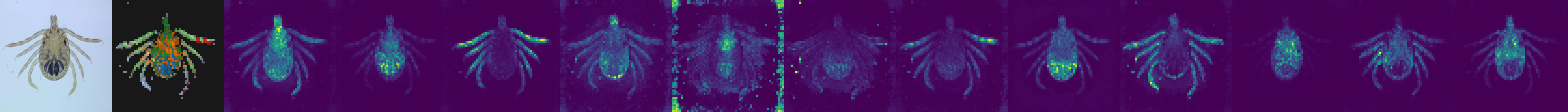}\\	
			\includegraphics[width=1.0\linewidth]{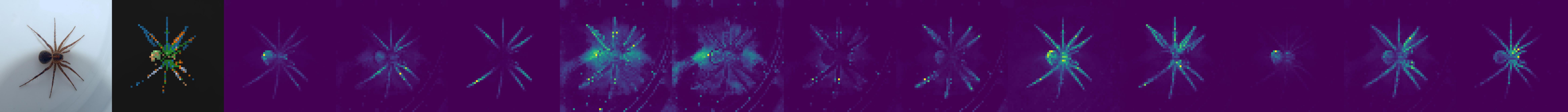}\\
		\end{tabular}
	\end{center}
	\vspace{-1.5em}
	\caption{Self-attention visualization of  ViT-B/16 pretrained by our Mugs. The images from left to right respectively denote the vanilla image, the overall self-attention score of all 12 heads in ViT-B, and the individual self-attention score of 12 heads.   \textbf{Best viewed in  color pdf file.} }
	\label{illustration_attention2}
\end{figure*}

\subsection{More Attention Visualization Results}\label{attention}
Here  same with  Fig.~\ref{illustration_attention} in the manuscript,  we  visualize more self-attention map of the 12 self-attention heads in ViT-B/16 pretrained by Mugs   in Fig.~\ref{illustration_attention2}.   The first column denotes the vanilla images, while each column of the last 12 columns  denote the self-attention score maps of each individual head.  The second column combines the 12 self-attention score maps from 12 heads into one, and also sets a threshold to remove some noises via only keeping top attention score. From these visualizations, one can observe that by using Mugs for pretraining, the overall self-attention of 12 heads   can capture the object shapes very well. For example, from the first bird image, it is even hard for human to get the bird location at the first glance, due to the similar color of the bird and the  flowers.  But the ViT-B/16 pretrained by Mugs still can well locate the bird and also capture the bird shape. Moreover, one  can also compare the attention visualization of Mugs with state-of-the-arts, \eg~iBOT. In iBOT~\cite{iBOT}, Fig.~18 in their appendix also visualizes the self-attention map. By comparison, the model pretrained by Mugs can better separate the object from background. These results testify that ViT-B/16 pretrained by Mugs can capture semantics in data even without any manual labels.

\begin{figure*}[tb]
	\begin{center}
		\setlength{\tabcolsep}{0.0pt}  
		\begin{tabular}{cccc c}
			\includegraphics[width=0.2\linewidth]{000000077595.jpg} & 
			\includegraphics[width=0.2\linewidth]{000000078823-A.jpg} & 
			\includegraphics[width=0.2\linewidth]{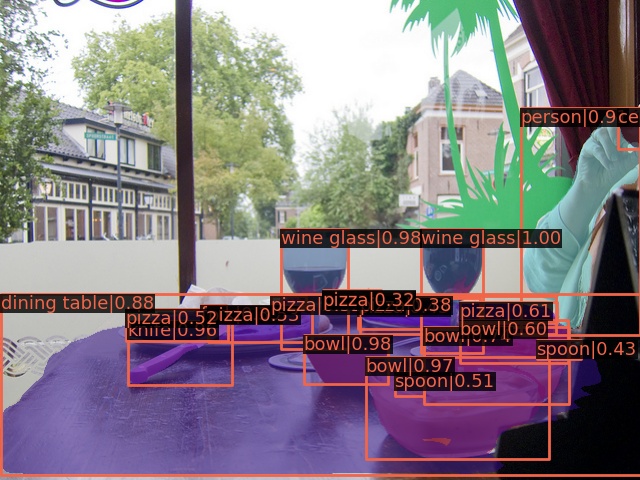}&
			\includegraphics[width=0.2\linewidth]{000000070254-A.jpg}& 
			\includegraphics[width=0.2\linewidth]{000000070048.jpg} \\
			\includegraphics[width=0.2\linewidth]{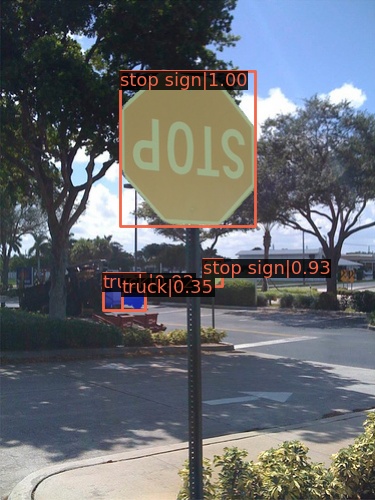} &
			\includegraphics[width=0.2\linewidth]{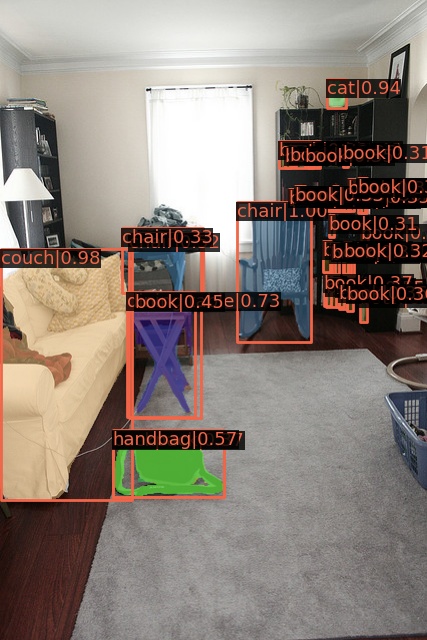}&
			\includegraphics[width=0.2\linewidth]{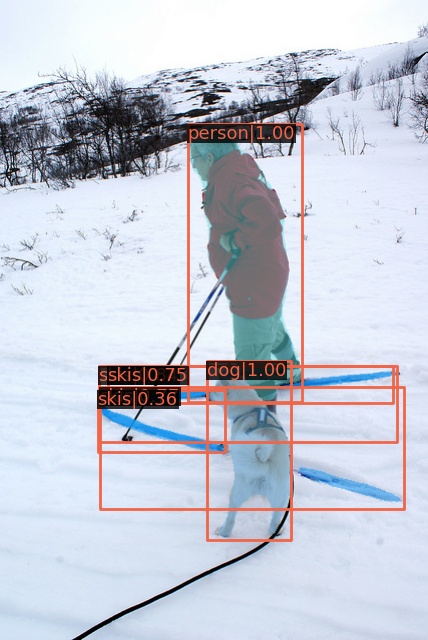} & 
			\includegraphics[width=0.2\linewidth]{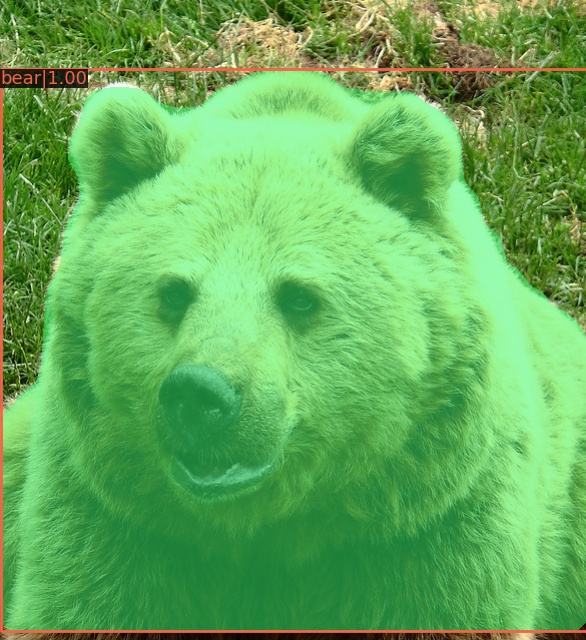} & 
			\includegraphics[width=0.2\linewidth]{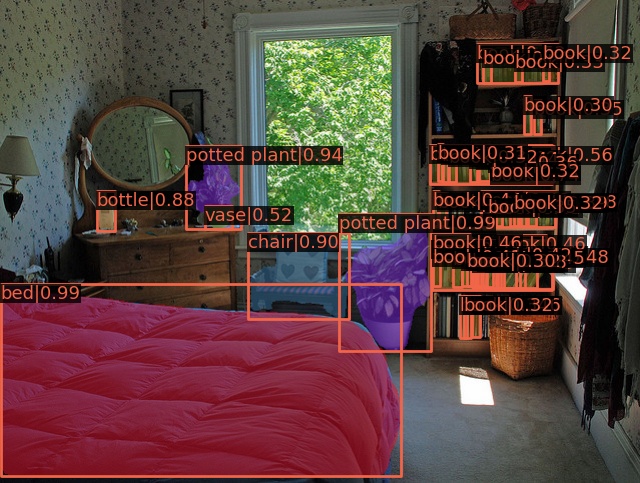}  \\
		\end{tabular}
	\end{center}
	\vspace{-1.5em}
	\caption{Object detection visualization   of ViT-B/16 pretrained by Mugs.   \textbf{Best viewed in  color pdf file.} }
	\label{illustration_detection}
\end{figure*}

\begin{figure*}[tb]
	\begin{center}
		\setlength{\tabcolsep}{0.0pt}  
		\begin{tabular}{cccc c}
			\includegraphics[width=0.2\linewidth]{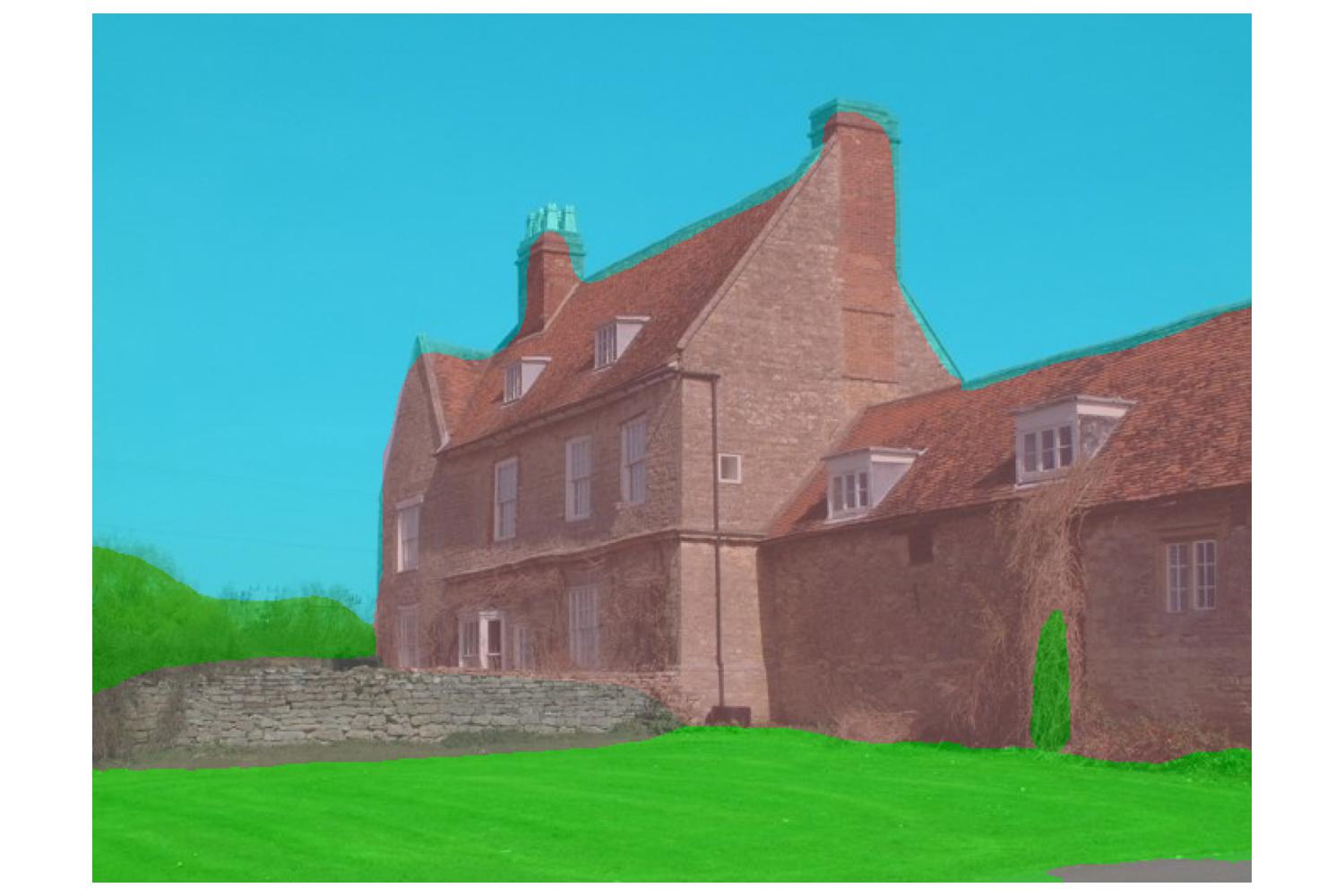} & 
			\includegraphics[width=0.2\linewidth]{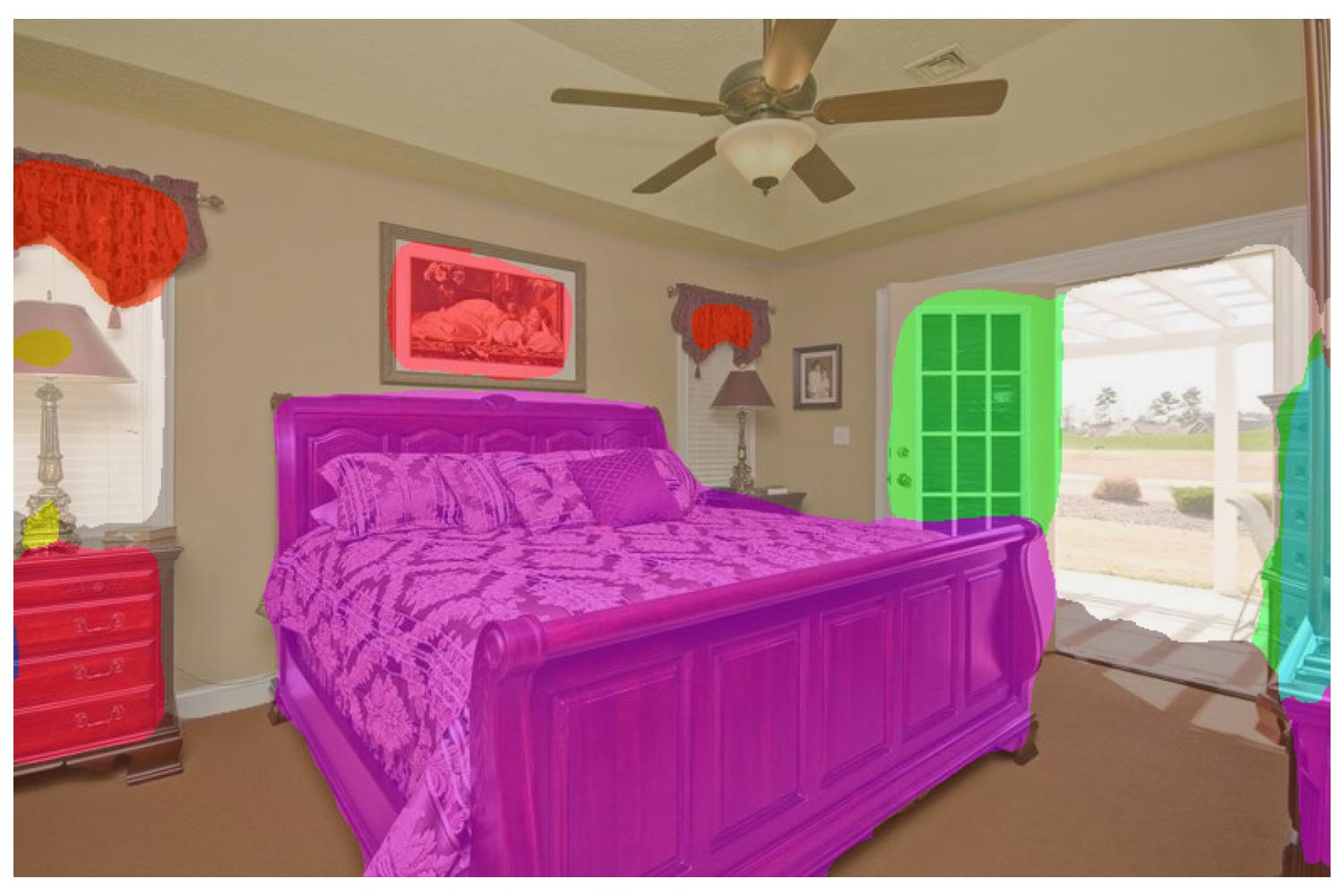} & 
			\includegraphics[width=0.2\linewidth]{ADE_val_00000129.jpg} & 
			\includegraphics[width=0.2\linewidth]{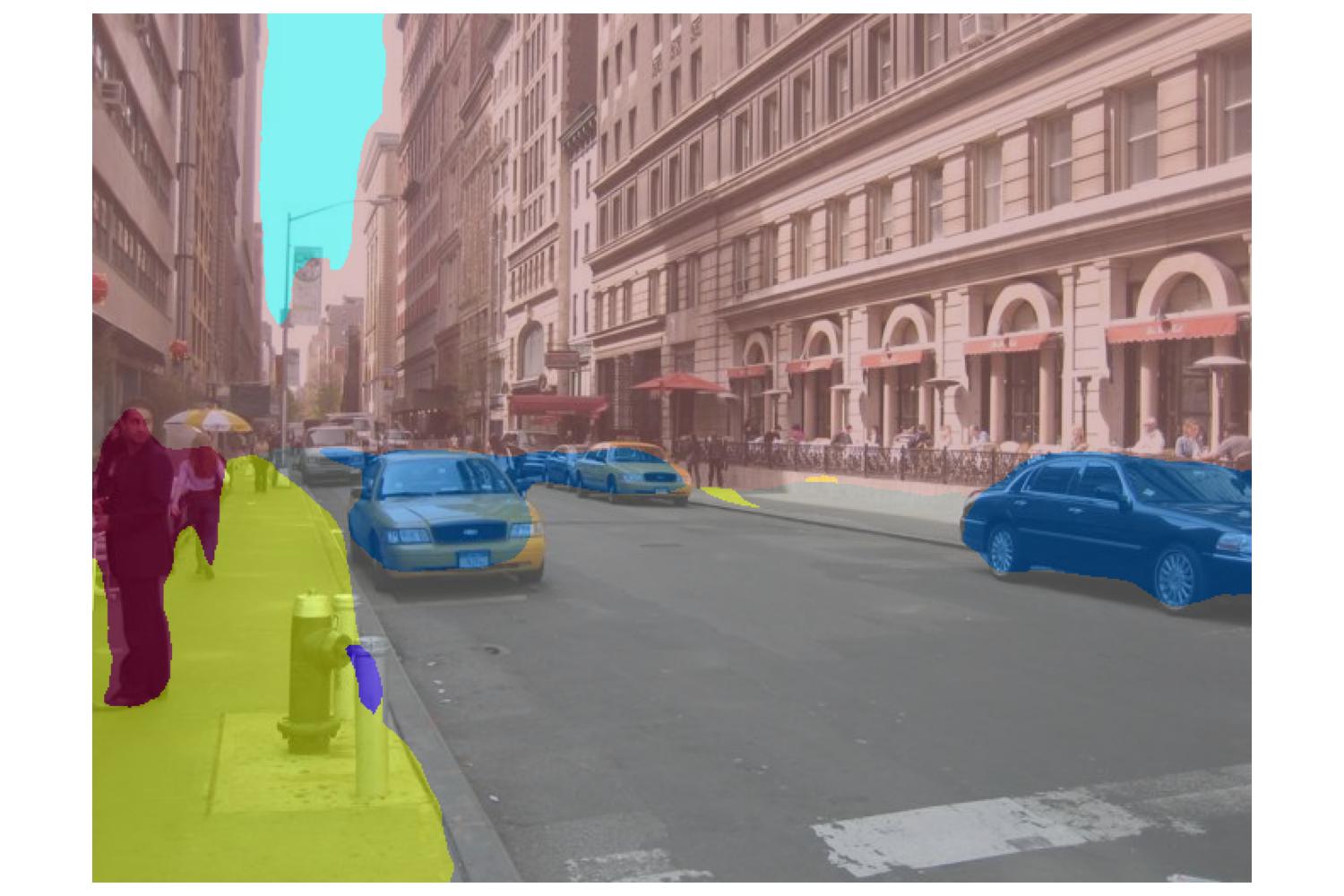} &
			\includegraphics[width=0.2\linewidth]{ADE_val_00000891.jpg} \\
			\includegraphics[width=0.2\linewidth]{ADE_val_00001233.jpg} &
			\includegraphics[width=0.2\linewidth]{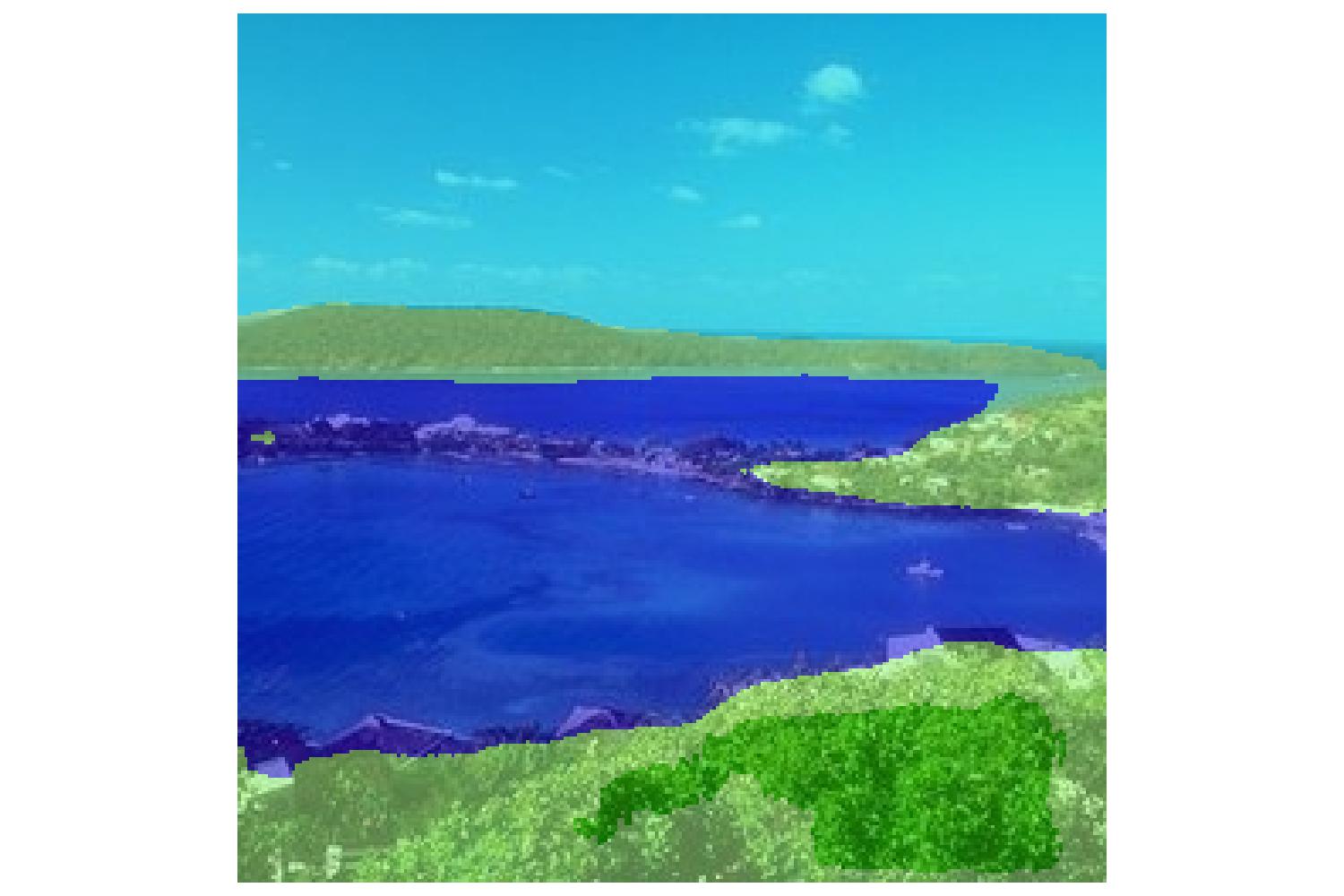} & 
			\includegraphics[width=0.2\linewidth]{ADE_val_00001859.jpg} &
			\includegraphics[width=0.2\linewidth]{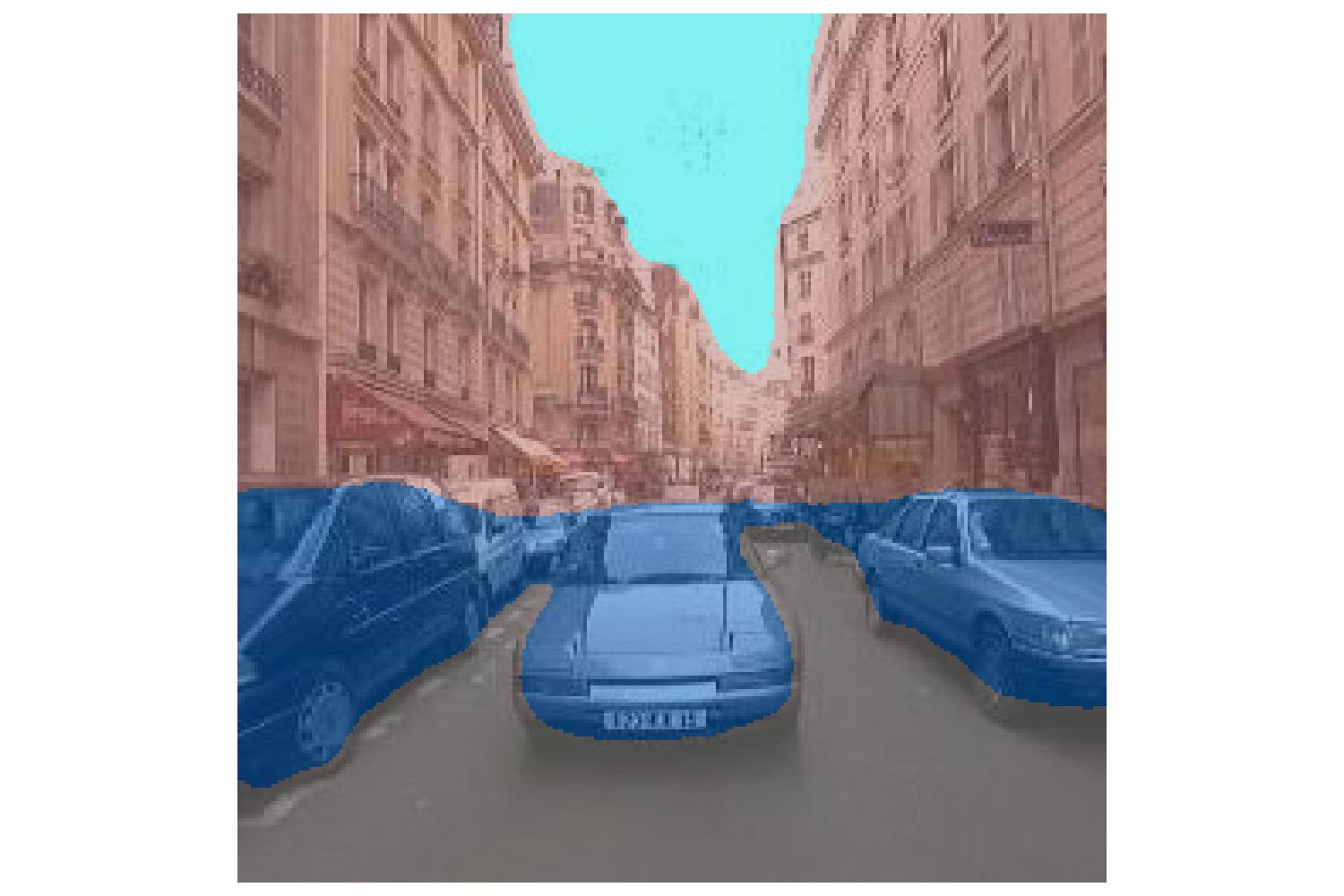} & 
			\includegraphics[width=0.2\linewidth]{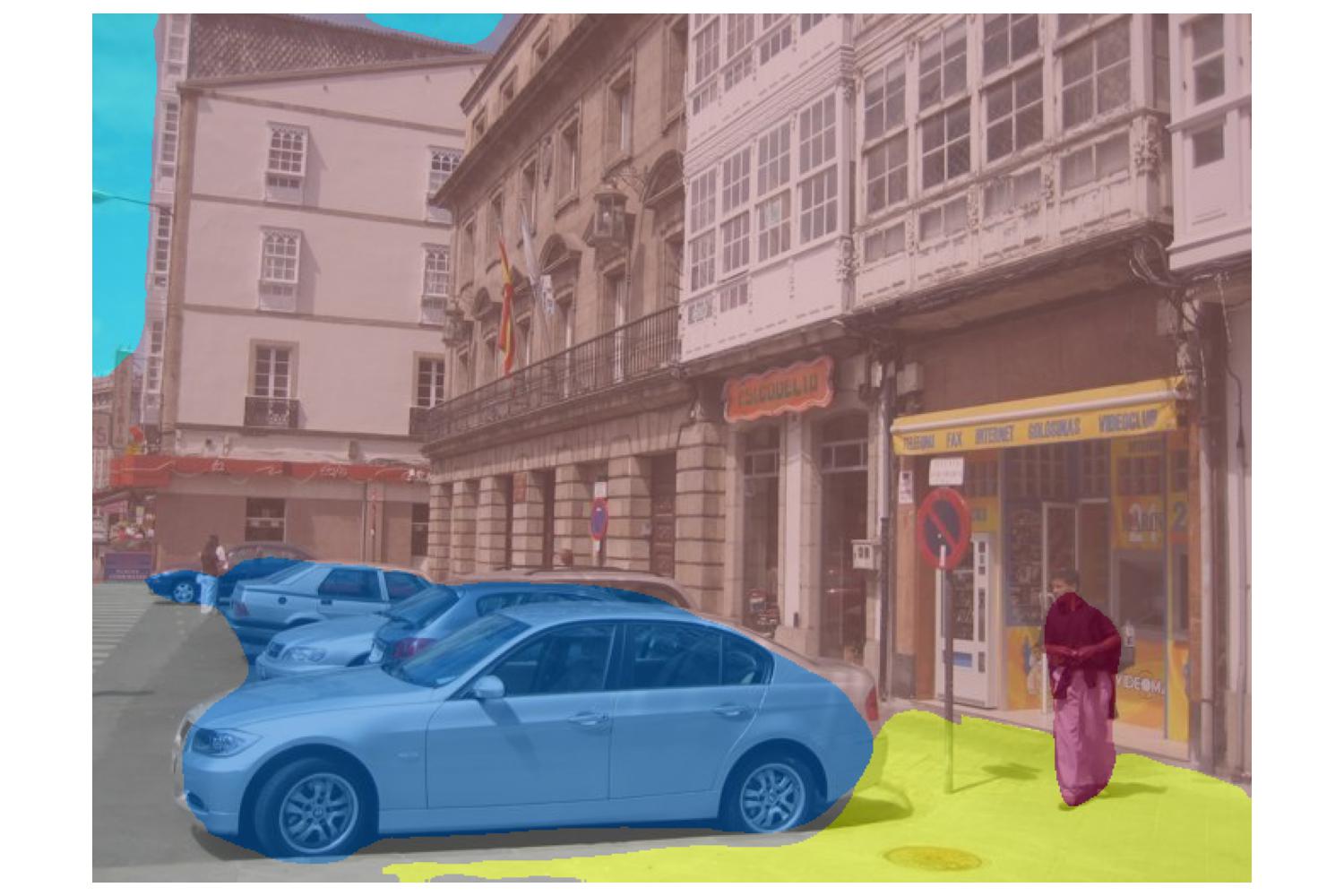}
		\end{tabular}
	\end{center}
	\vspace{-1.5em}
	\caption{Semantic segmentation visualization of ViT-B/16 pretrained by Mugs.   \textbf{Best viewed in  color pdf file.} }
	\label{illustration_segmentaiton}
\end{figure*}

\subsection{More Visualization Results on Object Detection and Semantic Segmentation}\label{Objectdetectionsec}
In the manuscript, we already provide some object detection and segmentation examples in Fig.~\ref{illustration_attention}. Here we give more examples.  Fig.~\ref{illustration_detection} shows more object detection examples on the COCO datasets, where we use the ViT-B/16 pretrained by our Mugs. From these results, one can observe that Mugs not only accurately  locate the objects in the images but also precisely recognizes these objects.   For semantic segmentation on ADE20K,  Fig.~\ref{illustration_segmentaiton} visualizes more examples. We also can find that Mugs can  capture the  object shape  accurately and thus well captures the semantics of an image.

\section{More Experimental details}\label{moreexp}
Due to space limitation, we defer more experimental  details to this section. Here we
first provide more details for pretraining. Then we present the experimental details for various kinds of downstream tasks. Finally, we also give the details for ablation study.

\subsection{More Details for Pretraining}\label{morepretraining}

\subsubsection{Augmentation.} 
Here the weak augmentation in our teacher backbone refers to the  augmentations used in many SSL works, \eg~DINO~\cite{DINO}, which includes random crop, color jitter, Gaussian noise, horizontal flipping and gray scaling. The hyper-parameters in these augmentation operations are the same with those in DINO~\cite{DINO}.  The strong  augmentation in our student backbone is the combination of  AutoAugment~\cite{AutoAugment} used  in DeiT~\cite{DeiT} and the above weak augmentation. Specifically, for each image, with probability 0.5,  we use AutoAugment~\cite{AutoAugment}  to augment it; otherwise,  we use weak augmentation to crop  it.    We use this sampling strategy to avoid training collapse while keeping sufficient data diversity. Following DINO and iBOT,   we always set the global crop scale as $[0.25, 1]$ and  local crop scale as $[0.05, 0.25].$

\subsubsection{Loss for multi-crop setting.}   Following conventional multi-crop setting~\cite{SwAV,DINO,iBOT}, we crop each image into 2 large crops of size $224$ and 10 extra small crops of size $96$. Then to construct the overall pretraining loss~\ref{overallloss} in manuscript, we regard one large crop as $\xmi{1}$ and respectively take the remaining 11 crops as $\xmi{2}$. Then symmetrically, we view another large crop as $\xmi{1}$ and respectively take the remaining 11 crops as $\xmi{2}$. Finally, we average these loss to obtain the overall training loss. 

\subsubsection{Hyper-parameter settings  for Pretraining.}   For all experiments, we use AdamW optimizer~\cite{AdamW}  with  a momentum of 0.9 and  a cosine learning rate schedule~\cite{SGDR}.  We also linearly warm up the learning rate at the first 10 epochs from $10^{-6}$ to its  base value, and then decay it with a cosine schedule~\cite{SGDR}.  For ViT-S and ViT-B, we  use a  minibatch size of 1024,  a base learning rate of $8\times10^{-4}$, and a weight decay of  0.1.   For ViT-L, due to  our limited computational resource,  we  use a  minibatch size of 640,  a base learning rate of $1.5\times10^{-4}$, and a weight decay of  0.08.   For all experiments,  the learning rate of the patch embedding layer is 5$\times$ smaller than the base learning rate.  This strategy is demontrated to be useful for stabling training in MoCo-v3~\cite{MoCo-v3}.  For drop path rate, we set it as 0.1/0.2/0.4 for ViT-S/B/L respectively. We set clip gradient as 3.0 for ViT-S/B and 0.3 for ViT-L. For Mugs, we follow MoCo to set $\tauin=\taulg=0.2$ in the infoNCE loss, and follow DINO to set $\tausg'=0.1$ and linearly warm up $\tausg$ from $0.04$ to $0.07$. We set the neighbor number $k=8$, and set  $\rho=0.9$  to estimate the center $\ptm$ 
in group discrimination.  All these settings are almost the same as DINO for simplicity which reduces hyper-parameter tuning and saves computational budget.

\subsubsection{Pretraining Cost.}  Mugs takes~about $27$  hours with $8$ A100  GPUs for  $100$ pretraining epochs on ViT-S/16. This means that Mugs has almost the same training cost with DINO, since  our projection/prediction heads and  transformers $\gtrans$  are much smaller than the backbone.  For ViT-B/16, Mugs needs about 24 hours on 16 A100 GPUs for $100$ pretraining epochs.  To training 100 epochs on ViT-L/16, Mugs takes about 48 hours on 40 A100 GPUs.  For ViT-B and ViT-L, it is hard for us to compare with DINO, since it does  not report the training time. 

\subsection{More Training Details for Evaluation  on ImageNet-1K} \label{1Kdetails}
%
%
%
%
%
%
%
\subsubsection{Fine-tuning.}  As mentioned in manuscript, we follow BEiT~\cite{BEiT}, DINO and iBOT, and use AdamW optimizer with layer-wise learning rate decay to train ViT-S/ViT-B/ViT-L for 200/100/50 epochs on ImageNet-1K. We  set   layer-wise learning rate decay  as 0.55 and learning rate $1.2\times 10^{-3}$  for both  ViT-S and ViT-B. For ViT-L,  we  use   layer-wise learning rate decay  0.75 and learning rate $8.0\times 10^{-4}$. For drop path rate, we set it as 0.1/0.2/0/3 for ViT-S/ ViT-B/ViT-L respectively. All these hyper-parameters are around at the suggested ones in BEiT and iBOT.

\subsubsection{Semi-supervised learning.} 
Following~DINO and iBOT,  we consider two settings: 1) training a logistic regression classifier on frozen features; and 2) fine-tuning the whole pretrained backbone. For logistic regression classifier, we  use AdamW optimizer with total minibatch size 1024 and weight decay 0.05, under both  1\% and 10\% training data settings.  We  sweep the learning rate $\{0.03, 0.06,$ $ 0.10, 0.2\}$. For fine-tuning 1000 epochs on ViT-S/16, we also use AdamW optimizer with total minibatch size 1024 and weight decay 0.05 under both  1\% and 10\% training data settings. We respectively  set learning rate as $2\times10^{-6}$ and $5\times10^{-6}$ for 1\% and 10\% training data.

%

\subsection{More Details for Downsteam Tasks} \label{downstreamtasks}

\subsubsection{Transfer learning.}   We  pretrain the model on ImageNet-1K, and then fine-tune the pretrained backbone on various kinds of other datasets with same protocols and optimization settings  in  DINO and iBOT. Specifically,  following DINO and iBOT, for both ViT-S and ViT-B, we always use AdamW optimizer with a minibatch size of 768. We fine-tune the pretrained model 360 epochs on  {{INat$_{18}$}} and {{INat$_{19}$}}, and 1000 epochs on {{Cif$_{10}$}}, {{Cif$_{100}$}}, {{Flwrs}} and {{Car}}.  For all datasets, we  sweep the learning rate $\{7.5\times 10^{-6}, 1.5\times 10^{-5}, 3.0\times 10^{-5}, 7.5\times 10^{-5}, 1.5\times 10^{-4}\}$.  For we set weight decay as $2\times 10^{-2}$ for CIFAR10 and CIFAR100 on ViT-B, and use a  weight decay of   $5\times 10^{-2}$ for all remaining experiments.  For example, on the {{INat$_{18}$}}  dataset, we use a learning rate of $3.0\times 10^{-5}$/$1.5\times 10^{-5}$ for ViT-S/ViT-B; on the {{INat$_{19}$}}  dataset, we set  learning rate as  $7.5\times 10^{-5}$/$3.0\times 10^{-5}$ for ViT-S/ViT-B. For more hyper-parameters, please refer to the hyper-parameter configure file in our released code. 









\subsubsection{Object detection \& Instance  segmentation.}   For fairness, we follow DINO and iBOT, and fine-tune the pretrained backbone via a  multi-scale strategy, namely resizing image at different scales. Please refer to iBOT for more details. We use AdamW optimizer with a learning rate of  $2\times 10^{-4}$, a weight decay of $0.05$ to  fine-tune with $1\times$ schedule, \ie~12 epochs with the learning rate decayed by $10\times$ at epochs $9$ and $11$.   
We sweep a layer decay rate of \{$0.65$, $0.75$, $0.8$, $0.9$\} and finally choose 0.8 because of its good performance.   For test, we do not use multi-scale strategy.

\subsubsection{Semantic segmentation.}
For semantic segmentation, we  follow DINO and iBOT, and fine-tune the pretrained backbone, and fine-tune the pretrained backbone by using $512\times512$-sized images for $1.6\times 10^4$ iterations. We use AdamW optimizer with a learning rate of  $2\times 10^{-4}$, a weight decay of $0.05$ and a layer decay rate of $0.9$  to  fine-tune.   For this task, we do not use any multi-scale strategy for  training and test.  
We sweep the learning rate $\{2\times10^{-5}, 3\times10^{-5}, 4\times10^{-5}, 5\times10^{-5}\}$ and finally choose $3\times10^{-5}$ because of its good performance.  


\subsection{ More details for Ablation Study} \label{appendixAblationStudy} 
Here we  provide the implementation details for  DINO and iBOT under different augmentations. 
As mentioned in Sec.~\ref{AblationStudy} in the manuscript, for the augmentation \scalebox{0.85}{$\Ts$}  in student network of  Mugs/DINO/iBOT, we implement it by  strong  or weak augmentation as mentioned at the beginning of Sec.~\ref{exp}; for augmentation \scalebox{0.85}{$\Tt$} in teacher, we always use weak augmentation. 

We first consider weak augmentation setting. For DINO,  there is no any change, since its vanilla version uses the weak augmentation. For Mugs, we only replace the strong augmentation used in the student network with the weak augmentation.  For iBOT, it has two losses, the proposed masked image modeling (MIM) loss and the clustering loss in DINO. To construct the MIM loss, iBOT needs to randomly mask the patches of input in the student network. But to build the clustering loss, it actually does not require random masks on input patches, but still uses the random masks in practice which actually increases the data augmentations for clustering loss. In this case, for fair comparison, we remove the random masks and only perform weak augmentation to construct the clustering loss in iBOT. Note, we still preserve the random masks for MIM loss to ensure MIM is the vanilla one in iBOT. 

We then consider strong augmentation setting.  For DINO,  we only replace the weak augmentation used in the student network with our strong augmentation.   For iBOT, same as the above for weak augmentation, we does not change the augmentation in MIM loss. But for building the clustering loss, we also remove the random masks and only perform strong augmentation.

\end{document}

%% file: MACPdef_manu.tex
\newcommand{\tauin}{\tau_{\mbox{\scriptsize{in}}}}
\newcommand{\taulg}{\tau_{\mbox{\scriptsize{lg}}}}
\newcommand{\tausg}{\tau_{\mbox{\scriptsize{g}}}}

\newcommand{\lamin}{\lambda_{\mbox{\scriptsize{in}}}}
\newcommand{\lamlg}{\lambda_{\mbox{\scriptsize{lg}}}}
\newcommand{\lamsg}{\lambda_{\mbox{\scriptsize{g}}}}

\newcommand{\hins}{h_{\mbox{\scriptsize{in}}}^{\mbox{\scriptsize{s}}}}
\newcommand{\hint}{h_{\mbox{\scriptsize{in}}}^{\mbox{\scriptsize{t}}}}
\newcommand{\pin}{p_{\mbox{\scriptsize{in}}}}

\newcommand{\hlgs}{h_{\mbox{\scriptsize{lg}}}^{\mbox{\scriptsize{s}}}}
\newcommand{\hlgt}{h_{\mbox{\scriptsize{lg}}}^{\mbox{\scriptsize{t}}}}
\newcommand{\plg}{p_{\mbox{\scriptsize{lg}}}}

\newcommand{\hsgs}{h_{\mbox{\scriptsize{g}}}^{\mbox{\scriptsize{s}}}}
\newcommand{\hsgt}{h_{\mbox{\scriptsize{g}}}^{\mbox{\scriptsize{t}}}}

\newcommand{\Bin}{\textsf{B}_{\mbox{\scriptsize{in}}}}
\newcommand{\Blg}{\textsf{B}_{\mbox{\scriptsize{lg}}}}

\newcommand{\gtrans}{g_{\mbox{\scriptsize{transformer}}}} 

\newcommand{\gt}{g_{\mbox{\scriptsize{transformer}}}^{\mbox{\scriptsize{t}}}} 
\newcommand{\gs}{g_{\mbox{\scriptsize{transformer}}}^{\mbox{\scriptsize{s}}}} 

\newcommand{\Ts}{\mathcal{\footnotesize{T}}^{\mbox{\scriptsize{s}}}}
\newcommand{\Tt}{\footnotesize{\mathcal{T}}^{\mbox{\scriptsize{t}}}} 

\newcommand{\ps}{\bm{p}^{\mbox{\scriptsize{s}}}} 
\newcommand{\pt}{\bm{p}^{\mbox{\scriptsize{t}}}} 
\newcommand{\ptm}{\bm{p}_{\mbox{\scriptsize{ema}}}}

\newcommand{\ymfi}[1]{\bm{y}_{#1}^{*}}

\newcommand{\ymsi}[1]{\bm{y}_{#1}'}

\newcommand{\xmi}[1]{\bm{x}_{#1}}

\newcommand{\xmti}[1]{\widetilde{\bm{x}}_{#1}}

\newcommand{\zmi}[1]{\bm{z}_{#1}}

\newcommand{\cmi}[1]{\bm{c}_{#1}}

\newcommand{\ymi}[1]{\bm{y}_{#1}}

\newcommand{\LL}{\mathcal{L}}
\newcommand{\J}{\mathcal{J}}

\newcommand{\rmw}[1]{\widetilde{\bm{r}}}
\newcommand{\rmb}[1]{\widehat{\bm{r}}}

\newcommand{\xmt}[1]{\widetilde{\bm{x}}}

\newcommand{\B}{\bm{\mathcal{B}}}

\newcommand{\F}{\mathcal{F}}